\PassOptionsToPackage{table,xcdraw}{xcolor}
\documentclass[sigconf, nonacm]{acmart}
\usepackage[table]{xcolor}

\AtBeginDocument{%
  }

\setcopyright{acmlicensed}
\copyrightyear{2018}
\acmYear{2018}
\acmDOI{XXXXXXX.XXXXXXX}

\acmConference[Conference acronym 'XX]{Make sure to enter the correct
  conference title from your rights confirmation emai}{June 03--05,
  2018}{Woodstock, NY}
\acmISBN{978-1-4503-XXXX-X/18/06}

\usepackage{float}
\usepackage{multirow}
\usepackage{tcolorbox}

\usepackage{subcaption}

\begin{document}

\title{UI-JEPA: Towards Active Perception of User Intent through Onscreen User Activity}

\author{Yicheng Fu}
\authornote{Work done as an intern at Apple.}
\authornote{Core Research Contributors: Yicheng implemented the UI-JEPA model, proposed the temporal masking strategy, fine-tuned and evaluated various video transformer and language model variants, and contributed to data processing and annotation. Ravi, as the technical lead, conceptualized UI-JEPA by integrating an SLM/LLM decoder with a JEPA encoder, identified key applications and processes for data collection, proposed the IIW and IIT datasets along with evaluation settings, researched and selected the SLM, LLM, and MLLM baselines, and was deeply involved in all stages of the project.}
\affiliation{
  \institution{Stanford University}
  \city{Stanford}
  \state{CA}
  \country{USA}
}

\author{Raviteja Anantha}
\authornote{Co-led datasets creation with equal contributions.}
\authornotemark[2]
\affiliation{
  \institution{Apple}
  \city{Seattle}
  \state{WA}
  \country{USA}
  }

\author{Prabal Vashisht}
\authornotemark[3]
\affiliation{
  \institution{Apple}
  \city{Seattle}
  \state{WA}
  \country{USA}
}

\author{Jianpeng Cheng}
\affiliation{
  \institution{Apple}
  \city{Seattle}
  \state{WA}
  \country{USA}
}

\author{Etai Littwin}
\affiliation{
  \institution{Apple}
  \city{Cupertino}
  \state{CA}
  \country{USA}
}


\begin{abstract}
Generating user intent from a sequence of user interface (UI) actions is a core challenge in comprehensive UI understanding.
Recent advancements in multimodal large language models (MLLMs) have led to substantial progress in this area, but their demands for extensive model parameters, computing power, and high latency makes them impractical for scenarios requiring lightweight, on-device solutions with low latency or heightened privacy.
Additionally, the lack of high-quality datasets has hindered the development of such lightweight models.
To address these challenges, we propose \emph{UI-JEPA}, a novel framework that employs masking strategies to learn abstract UI embeddings from unlabeled data through self-supervised learning, combined with an LLM decoder fine-tuned for user intent prediction. 
We also introduce two new UI-grounded multimodal datasets, ``Intent in the Wild'' (IIW) and ``Intent in the Tame'' (IIT), designed for few-shot and zero-shot UI understanding tasks. 
IIW consists of 1.7K videos across 219 intent categories, while IIT contains $\sim$900 videos across 10 categories.
We establish the first baselines for these datasets, showing that representations learned using a JEPA-style objective, combined with an LLM decoder, can achieve user intent predictions that match the performance of state-of-the-art large MLLMs, but with significantly reduced annotation and deployment resources. 
Measured by intent similarity scores, UI-JEPA outperforms GPT-4 Turbo and Claude 3.5 Sonnet by 10.0\% and 7.2\% respectively, averaged across two datasets.
Notably, UI-JEPA accomplishes the performance with a 50.5x reduction in computational cost and a 6.6x improvement in latency in the IIW dataset. These results underscore the effectiveness of UI-JEPA, highlighting its potential for lightweight, high-performance UI understanding.
\end{abstract}





\maketitle


\section{Introduction}
\label{sec:intro}
As the use of smart devices for daily tasks increases, the user interface (UI) remains the primary medium through which humans interact with applications, either directly or via dialogue agents. Accurately perceiving UI actions and predicting user intent can significantly enhance dialog agents, providing valuable feedback on the success or failure of their actions. Moreover, an effective perception model can enable ``Multimodal Intent State Tracking'' (MIST) that summarizes user interaction history with a smart device.
However, understanding UIs poses significant challenges. Applications vary in visual representation, which can change dynamically based on user actions. This necessitates using cross-modal features — images, natural language, and structural metadata — to grasp the temporal relationships in UI sequences.

Recent advances in Multimodal Large Language Models (MLLMs) have made progress in building perception agents capable of generating user intent from UI action sequences. However, most state-of-the-art MLLMs~\cite{openai_2024_gpt4, anthropic_2024_claude} are computationally intensive and require server-side processing, leading to extremely high costs. Given connectivity, latency, and privacy concerns, an on-device solution is needed — one that is lightweight yet maintains comparable accuracy. While efforts are underway to reduce the size of these models~\cite{openai_2024_gpt4omini}, they are still far from the optimal size ($\sim$3B parameters) needed for reliable operation on advanced mobile devices. Additionally, these models still require substantial data and compute resources for training.

Inspired by the success of self-supervised learning (SSL) techniques like Joint Embedding Predictive Architectures (JEPA)~\cite{lecun_2022_jepa} and its variants~\cite{assran_2023_ijepa, bardes_2024_vjepa}, we propose UI-JEPA, a lightweight video-to-text model specialized for UI activities. UI-JEPA comprises a JEPA-based encoder, trained with novel temporal masking strategies on unlabeled UI video data, to learn abstract feature representations, and an LLM decoder that predicts user intent from these features. Our key insight, inspired by the \textit{Predictive Feature Principle}~\cite{rao1999predictivefeatureprinciple}, is that predicting fully masked frames using unmasked frames allows the model to effectively capture temporal relationships and understand task meanings.
We demonstrate that fine-tuning an LLM decoder conditioned on UI-JEPA representations requires only a fraction of the paired video-text data and computational resources as required by state-of-the-art MLLMs. This framework is particularly valuable when high-quality labeled data are scarce.

\begin{figure*}[ht!]
    \centering
    \begin{subfigure}[t]{0.5\textwidth}
        \centering
        \includegraphics[width=\textwidth]{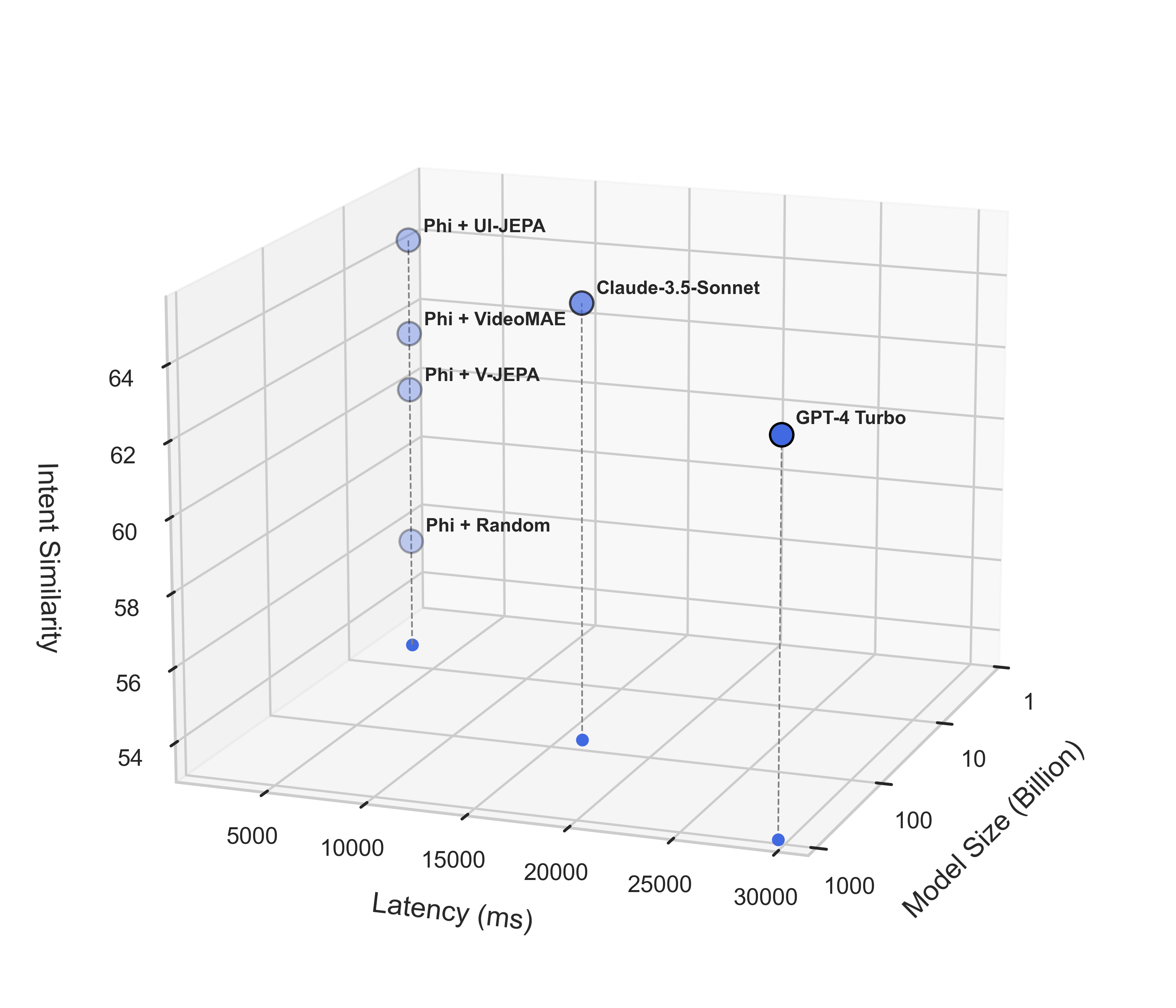}
        \caption{Intent in the Wild Dataset.}
    \end{subfigure}%
    \hfill
    \begin{subfigure}[t]{0.5\textwidth}
        \centering
        \includegraphics[width=\textwidth]{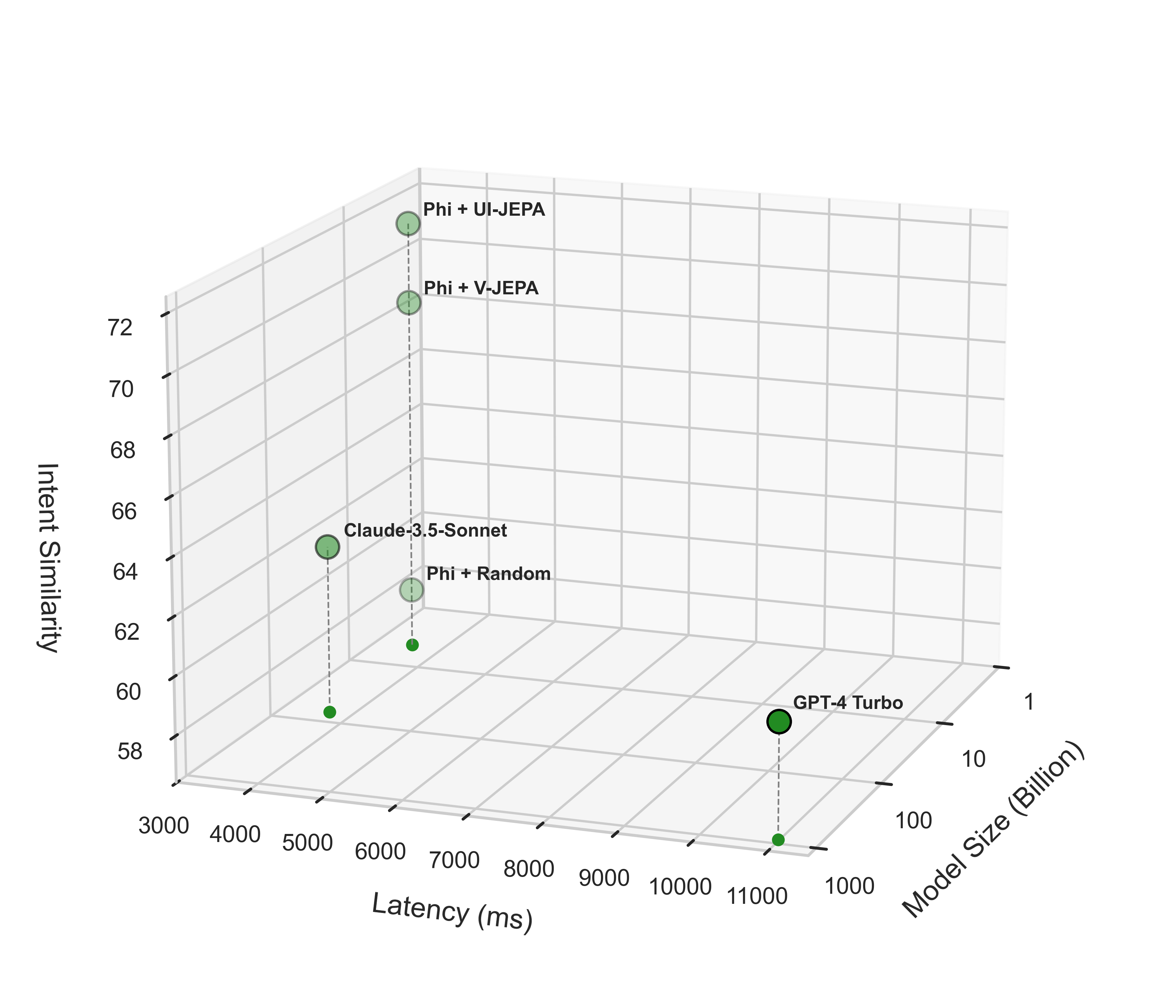}
        \caption{Intent in the Tame Dataset.}
    \end{subfigure}
    \caption{3D Scatter Plots Comparing Benchmark Scores with Model Size and Latency in Intent in the Wild and Intent in the Tame dataset respectively: (a) the relationship between model size (in billions of parameters), latency (in milliseconds), and Intent similarity scores; (b) the same relationship but for Intent in the Tame dataset. Each point represents a different model.}
    \label{fig:benchmark_3d}
\end{figure*}


The lack of high-quality, task-specific labeled UI datasets has also hindered the development of lightweight MLLMs for UI understanding. To address this, we introduce two new multimodal datasets: ``Intent in the Wild'' and ``Intent in the Tame.'' 
We establish the first benchmarks on these datasets using the UI-JEPA framework in both few-shot and zero-shot settings. 
Our contributions are:
\vspace{-1mm}
\begin{itemize}
\item  {\textbf{Benchmarks}}: We introduce two new benchmarks, ``Intent in the Wild'' and ``Intent in the Tame,'' both evaluated in few-shot and zero-shot settings for UI understanding. The task involves generating user intent from a sequence of UI actions captured in video format.\footnote{The datasets will be made publicly available shortly.}
\item  {\textbf{Framework}}: We propose UI-JEPA, a novel framework that employs various masking strategies to learn abstract UI embeddings from unlabeled data through SSL.
\item  {\textbf{Model}}: We present a lightweight JEPA-tuned MLLM (JEPA-based video encoder + fine-tuned auto-regressive head) designed to generate user intent from UI action sequences, showcasing the integration of JEPA with LLM decoders for user intent prediction.
\item  {\textbf{Comparison}}: We compare our lightweight model against state-of-the-art MLLMs~\cite{openai_2024_gpt4, anthropic_2024_claude}, demonstrating that our approach achieves parity while using only a fraction of the data, time, and computational resources. See Figure~\ref{fig:benchmark_3d} for a comparison.\footnote{Estimated parameter count based on publicly available information\cite{thompson2024claude3}.}
\end{itemize}

\section{Related Work}
\label{sec:rel_work}
Our work with UI-JEPA is focused on advancing UI understanding by significantly improving generalization capabilities and enabling automatic inference of user intent from UI interaction sequences with minimal reliance on annotation, memory, and computational resources during both training and inference. By using JEPA, we strive to achieve performance matching large state-of-the-art MLLMs while utilizing less video-text aligned training data. This work sits at the intersection of UI understanding, MLLMs, and self-supervised learning, pushing the capabilities of what is possible with lightweight MLLMs.

\begin{figure*}[ht!]
    \centering
    \begin{subfigure}[t]{0.7\textwidth}
        \centering
        \includegraphics[width=\textwidth]{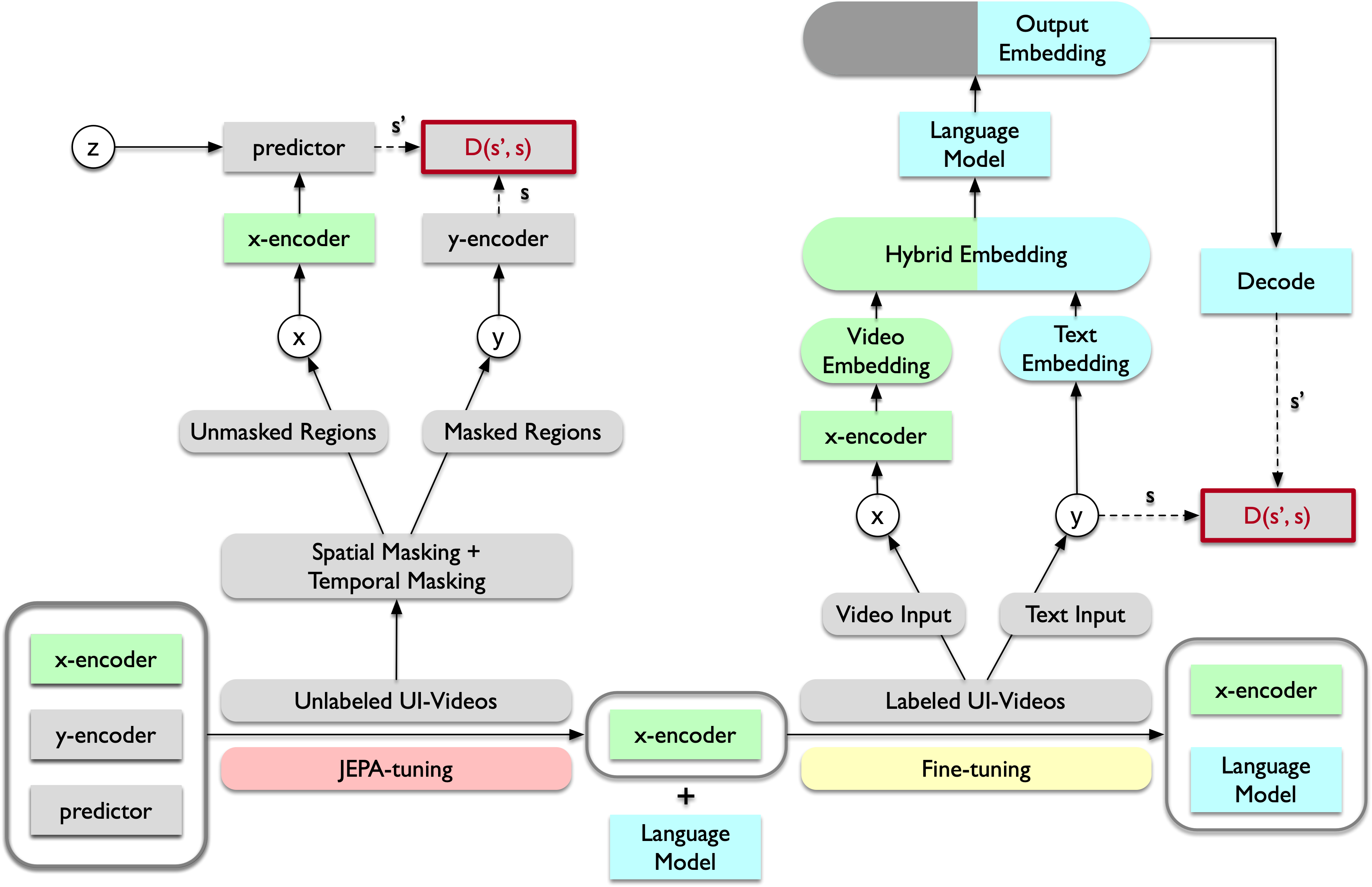}
        \caption{Training Process.}
    \end{subfigure}%
    \hfill
    \begin{subfigure}[t]{0.25\textwidth}
        \centering
        \includegraphics[width=\textwidth]{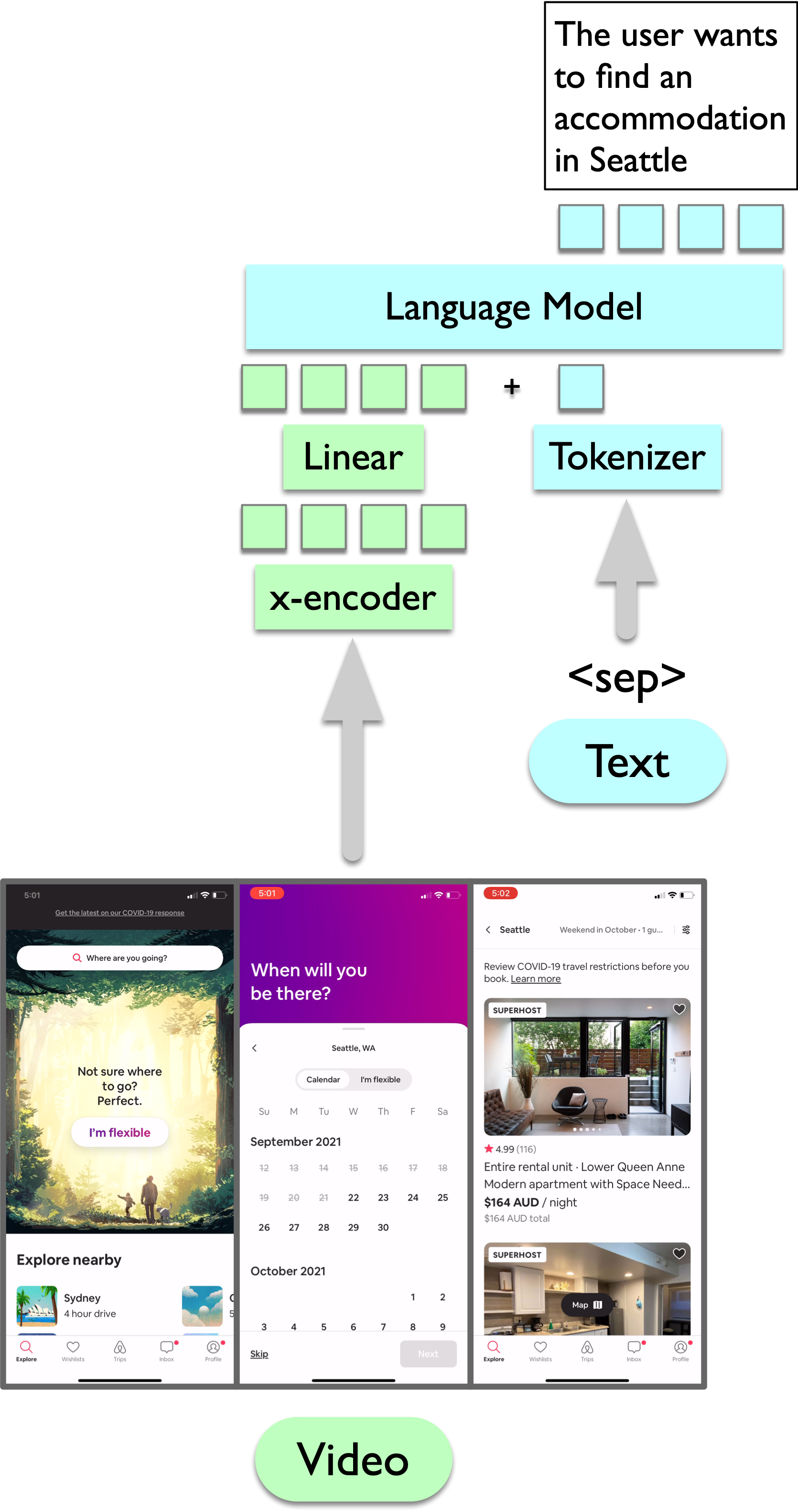}
        \caption{Inference Process.}
    \end{subfigure}
    \caption{(a) Training Process of UI-JEPA: The training process consists of two stages: (1) JEPA tuning Stage: The pre-trained x-encoder, y-encoder, and predictor are further fine-tuned on our UI datasets using various masking techniques. (2) LLM Fine-tuning Stage: The parameters of the x-encoder from the previous stage is frozen. The video embedding is combined with text tokens embeddings, and fed together as inputs to the large language model to generate an output embedding. The final loss is computed based only on the text portion of the output, excluding the video portion; (b) Inference Process of UI-JEPA: During inference, the video embedding and text embeddings are input into the language model to generate a prediction of user intent.}
    \label{fig:process}
\end{figure*}


\subsection{UI Understanding}
Various machine learning models have been proposed to enhance UI understanding. Early efforts~\cite{he-2021-actionbert, bai-2021-uibert} primarily focused on pre-training transformer-based models using large-scale, unlabeled UI data to learn generic feature representations at the UI component level. Other approaches~\cite{zhang-2021-screenrecognition} have augmented model training with semantic information and heuristics to improve UI detection. However, these methods often fall short in understanding the concept of a task and fail to learn comprehensive visual representations, as they are limited to individual UI components. Some approaches~\cite{wu-2023-neverendinglearning} utilize crawler models tailored to specific tasks, but these models struggle with scalability across a large number of tasks and exhibit limited generalization to unseen tasks. Additionally, methods~\cite{you2024ferretuigroundedmobileui} that integrate image encoders with LLMs are generally confined to basic UI tasks, such as icon recognition and widget listing, and operate on static images, which hinders their ability to learn temporal relationships and the concept of a task.

In contrast, our approach processes videos that capture sequences of UI actions during task execution. By using a JEPA-based encoder, we learn video representations through self-supervised learning, and an LLM decoder to textual representations of the user intents. This method captures not only the temporal dynamics of UI interactions, but also offers a more holistic understanding of user tasks.

\subsection{Multimodal Large Language Models}
Recent Multimodal Large Language Models~\cite{openai_2024_gpt4, anthropic_2024_claude} (MLLMs) have led to significant advancements in UI understanding, combining various data modalities for more precise intent predictions. Despite these improvements, the high computational demands of MLLMs often necessitate server-side processing, which drives up costs and limits their scalability, while potentially sacrificing privacy and imposing connectivity constraints.

Recent research~\cite{openai_2024_gpt4omini} has focused on reducing model size while striving to maintain performance. Although these efforts have achieved some success, the resulting models still require substantial data and remain too large for efficient deployment on advanced mobile devices.

In comparison, our approach seeks to address these limitations by further compressing the model size and reducing data requirements. We achieve this without compromising performance, demonstrating that our method performs competitively with larger models in both few-shot and zero-shot scenarios, making it more suitable for mobile environments.

\begin{figure*}[ht!]
    \centering
    \begin{minipage}{0.45\textwidth}
        \centering
        \includegraphics[width=\textwidth]{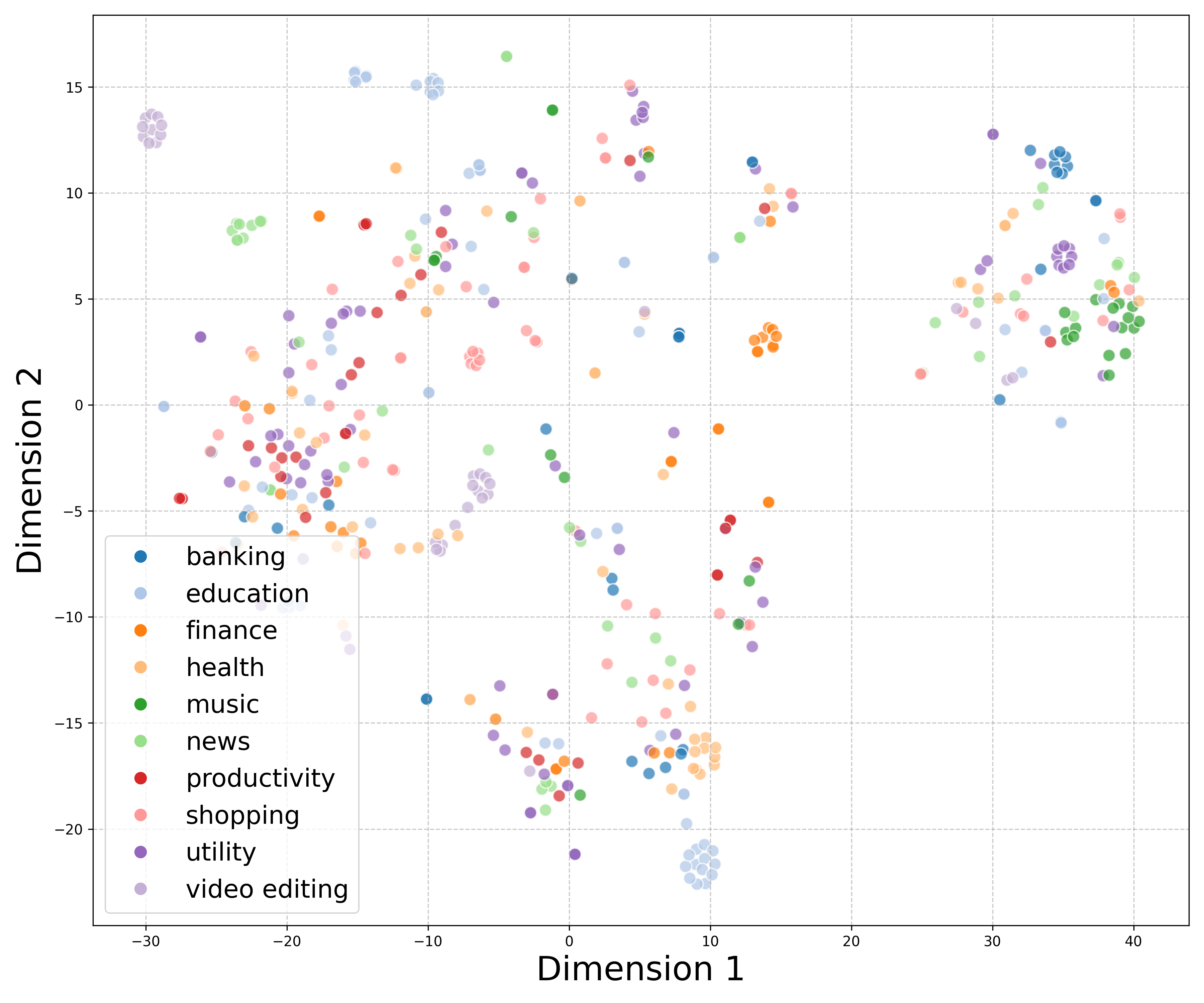}
    \end{minipage}\hfill
    \begin{minipage}{0.45\textwidth}
        \centering
        \includegraphics[width=\textwidth]{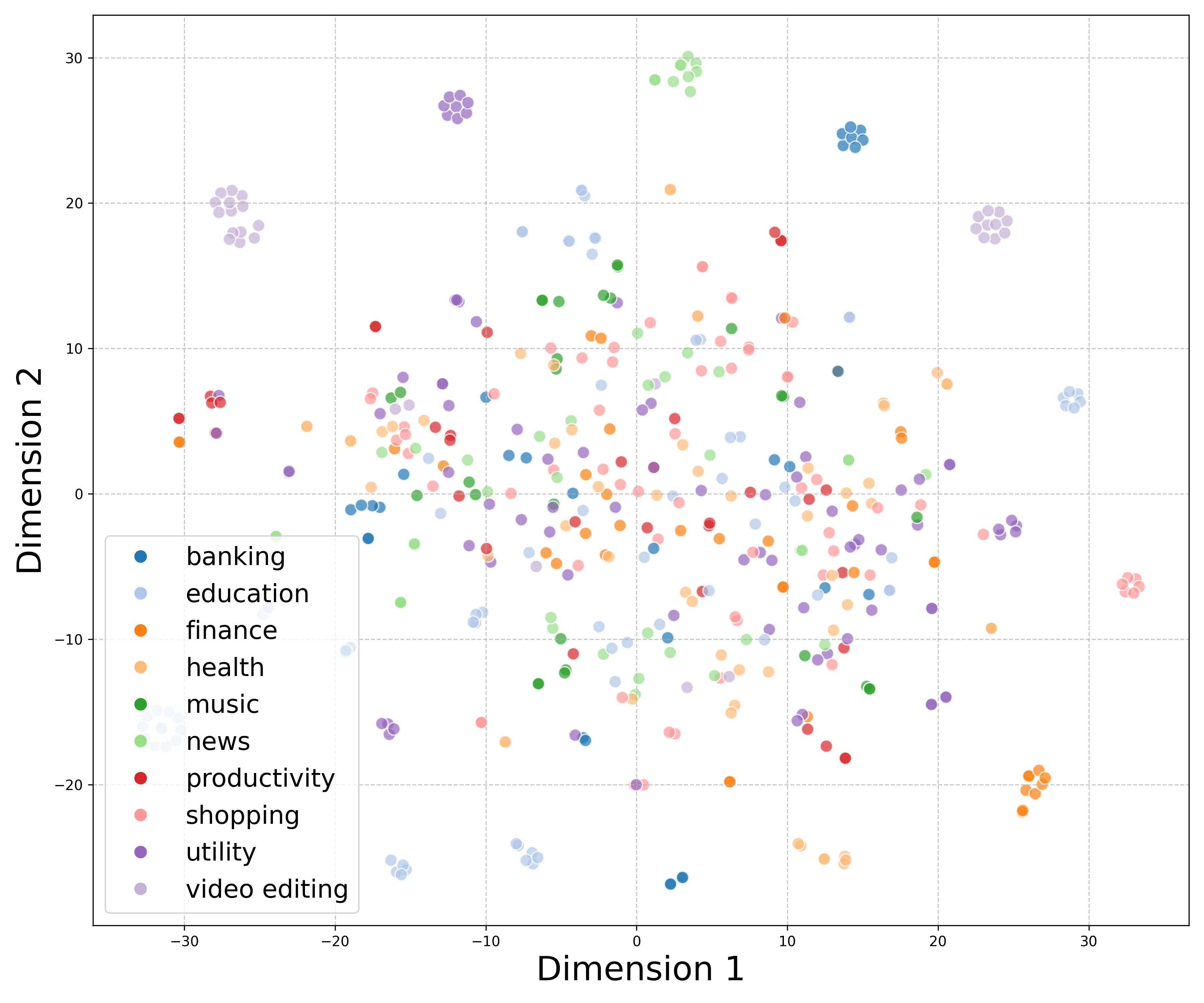}
    \end{minipage}
    \caption{2D Visualization of Video Embeddings: The left panel shows the 2D embedding representation of videos from the ``Intent in the Wild'' dataset using a random encoder, while the right panel displays the embeddings generated by the UI-JEPA encoder.}
    \label{fig:video_embedding}
\end{figure*}

\subsection{Self Supervised Learning}
Obtaining large-scale labeled datasets for UI understanding, especially those that pair videos of UI actions with user intent labels, is costly and difficult to scale. Moreover, such datasets cannot capture every possible visual variation, making it essential to develop approaches that use unlabeled data and generalize well by learning robust abstract representations.

Self-supervised learning (SSL) has emerged as a promising solution to these challenges. For instance, VideoMAE~\cite{tong2022videomaemaskedautoencodersdataefficient} uses SSL to pretrain vision transformers~\cite{dosovitskiy2021imageworth16x16words} (ViTs) by masking and reconstructing random video cubes. Similarly, Joint Embedding Predictive Architecture~\cite{lecun_2022_jepa} (JEPA) and its derivatives, I-JEPA~\cite{assran_2023_ijepa} and V-JEPA~\cite{bardes_2024_vjepa}, focus on learning semantic representations by ignoring irrelevant details and predicting masked spatio-temporal regions.

Our approach builds on existing JEPA-based methods by employing a temporal masking strategy, where entire frames are masked rather than just patches. Additionally, we integrate a JEPA-based encoder with an LLM decoder, projecting video embeddings from ViT into the LLM input space using a dense layer and generating intent descriptions with a fine-tuned LoRA~\cite{hu2022lora} adapter.

\section{The \emph{UI-JEPA} Framework}
\label{sec:framework}

Our goal is to track a mobile user's intent by analyzing their interactions with the UI and representing that user intent as a text summary. We opt for text over a structured format because natural language serves as a general-purpose and scalable semantic representation, which language models excel at generating.

This task requires comprehending the textual, spatial, and temporal dimensions of screen activities and transforming their abstract meanings into a coherent text description. To address this, we developed UI-JEPA, a framework consisting of two key components: a video transformer and a decoder-only language model (LM). The video transformer processes videos of continuous screen activities into video embeddings, which are then fed into the decoder-only LM to generate a corresponding text description of the user intent.

\subsection{Network Parameterization}
We employ the Vision Transformer~\cite{dosovitskiy2021imageworth16x16words} (ViT) as our video encoder. To obtain the video embeddings, we process the entire video by sampling 16 evenly spaced frames, resizing them to 384 $\times$ 384 pixels, and dividing them into a 3D grid of spatial-temporal patches. Each patch measures 16 $\times$ 16 pixels and spans 2 frames. These patches, or visual tokens, are fed into the video encoder to generate the video embeddings. As illustrated in Figure~\ref{fig:process}(b), these embeddings are projected into the LM's input space using a dense layer.

For the LM, we use a lightweight variant, Microsoft Phi-3~\cite{abdin2024phi}, with approximately 3 billion parameters. This choice of a lightweight model facilitates on-device experimentation and deployment. The LM processes both the video embeddings and the text embeddings, with positional embeddings applied only to the text inputs. An overview of UI-JEPA inference architecture is presented in Figure~\ref{fig:process}(b).

\begin{figure*}[ht!]
    \centering
    \includegraphics[width=\textwidth]{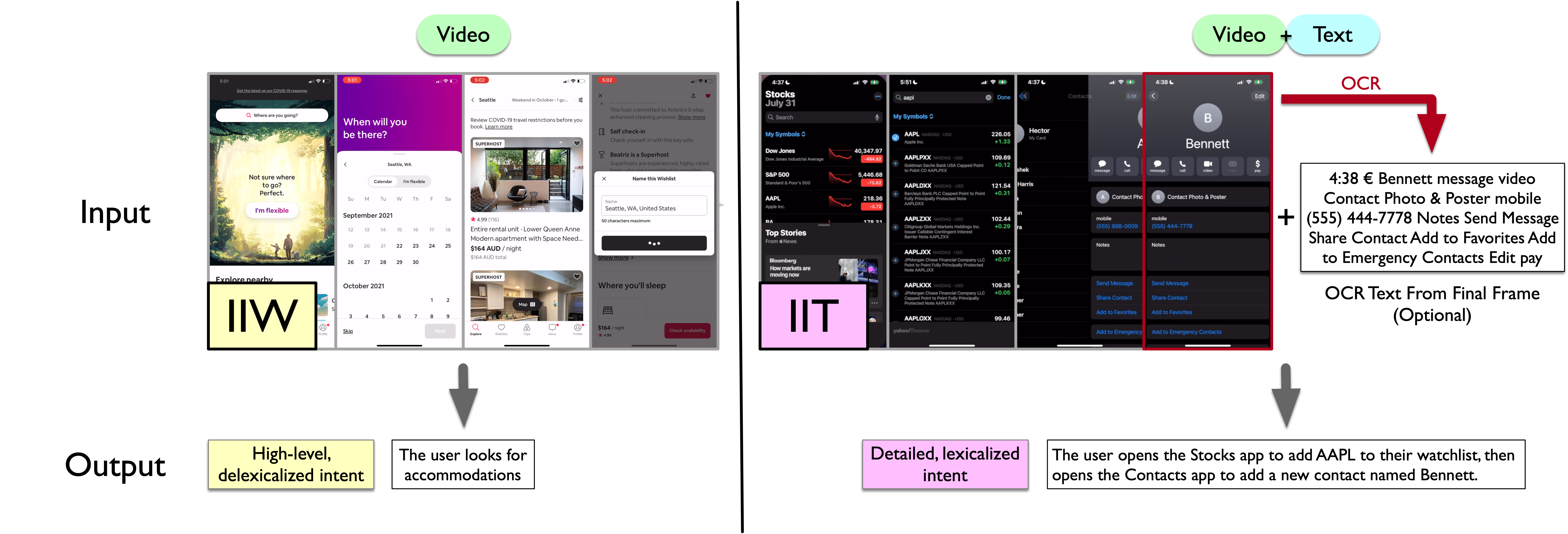}
    \caption{Examples of inputs and corresponding labels from the IIW and IIT datasets. In the IIW dataset, the input is a sequence of UI actions in a single video, labeled with a high-level, delexicalized description of user intent. In contrast, the IIT dataset uses lexicalized intent as the label. In addition, it includes OCR texts converted from the final video frame.}
    \label{fig:data_example}
\end{figure*}

\subsection{Training}
We use pre-trained weights from a Vision Transformer (ViT) and an LM to perform fine-tuning in two stages: first, fine-tuning the ViT on unlabeled UI videos, and second, fine-tuning the LM on videos labeled with user intent descriptions.

In the initial fine-tuning stage for the ViT, we address the challenge of annotating user intent descriptions by training the ViT on solely videos in a self-supervised manner. Following the approach of V-JEPA~\cite{bardes_2024_vjepa}, we use a masked prediction task where the unmasked video serves as context, and the masked part is predicted as the target. As illustrated in Figure~($x$-encoder), a target momentum encoder ($y$-encoder), and a predictor. The context encoder uses the ViT weights, which are the focus of our fine-tuning. During JEPA fine-tuning, we apply a masking strategy: masked tokens are removed from the input to the $x$-encoder, but are used as inputs to the $y$-encoder, along with the unmasked tokens (we discuss the different masking schemes employed in~\ref{subsec:abalation}). The predictor receives a sequence of embeddings from the $x$-encoder concatenated with learnable mask tokens, which include positional embeddings representing the 3D spatio-temporal positions of the masked tokens. The predictor is then tasked with predicting the $y$-encoder's embeddings for each mask token. The $y$-encoder weights start as a deep copy of the $x$-encoder weights and are updated using an exponential moving average (EMA)~\cite{tarvainen2018meanteachersbetterrole} of the $x$-encoder weights.

In the second fine-tuning stage, we freeze the ViT weights (the $x$-encoder from UI-JEPA) and update the adapters of a pre-trained language model. The video embeddings produced by the ViT are projected into the LM's input space via a dense layer. The parameters updated in this stage include the dense projection layer and the LM's LoRA adapter~\cite{hu2022lora, dettmers2023qloraefficientfinetuningquantized}. An overview of the training process is shown in Figure~\ref{fig:process}(a).

\subsection{UI-JEPA Data Strategy}

There are two key differences in the fine-tuning data between UI-JEPA and its parent V-JEPA. First, we intentionally avoid data augmentation on UI videos. Unlike general video data, UI videos contain both spatial and temporal activities as well as textual information that accurately describes app functionality and user input. Our experiments demonstrate that random data augmentation can result in malformed UI videos, which negatively impacts model performance. Second, while UI-JEPA incorporates the patch-based temporal-spatial masking strategy used in V-JEPA, it also introduces a novel temporal masking approach where entire frames are masked rather than just patches. This new strategy enhances the model's ability to learn frame dependencies, addressing the dramatic changes that often occur in UI videos when users open new apps or navigate between different screens.

\subsection{Visualization of UI-JEPA embeddings}
To evaluate the effectiveness of the UI-JEPA encoder in extracting abstract representations, we first select the top 10 most frequent app types from the IIW dataset and created a 2D embedding visualization using the t-SNE method~\cite{van2008visualizing}. We compare this representation to that of a randomly initialized video encoder, as illustrated in Figure~\ref{fig:video_embedding}. To assess cluster quality, we compute the silhouette score, which measures how closely objects in a cluster resemble each other compared to objects in other clusters. As shown in Figure~\ref{fig:video_embedding}, UI-JEPA achieves higher silhouette scores than baselines, indicating that the UI-JEPA embeddings are more tightly packed within their clusters. 

Additionally, we calculate the cosine similarity for each pair of video embeddings and compare these to the cosine similarity scores of corresponding natural language intent pairs. This analysis investigates the correlation between these similarity scores, with the expectation that videos with similar intents will have similar embeddings. A higher correlation suggests that our video encoder effectively captures high-level representations of user intent. As shown in Table~\ref{tab:encoder_performance},  UI-JEPA leads to higher video-text correlation scores compared to V-JEPA and other baselines introduced in section~\ref{baselines}.

\begin{table}
  \caption{Performance Metrics for Different Encoders: P-correlation (Pearson correlation coefficient), S-correlation (Spearman's rank correlation coefficient), and S-Score (Silhouette Score). These metrics evaluate the performance of various encoders.}
  \label{tab:encoder_performance}
  \begin{tabular}{lccc}
    \toprule
    \textbf{Encoder} & \textbf{P-Correlation} & \textbf{S-Correlation} & \textbf{S-Score} \\
    \midrule
    Random & 0.0334 & 0.0155 & -0.1230 \\
    VideoMAE & 0.0861 & 0.0291 & 0.0081 \\
    V-JEPA & 0.1158 & 0.0435 & 0.0094 \\
    UI-JEPA & 0.1267 & 0.0427 & 0.0212 \\
    \bottomrule
  \end{tabular}
\end{table}

\begin{table*}
  \caption{Data statistics.}
  \label{tab:fata_statistics}
  \begin{tabular}{ccccccc}
    \toprule
    \multirow{2}{*}{\textbf{Dataset Name}} & \multirow{2}{*}{\textbf{Resolution}} & \multicolumn{3}{c}{\textbf{Number of Videos}} & \multirow{2}{*}{\textbf{Avg. Frames}} & \multirow{2}{*}{\textbf{Categories}} \\
    \cmidrule(l){3-5}
    & & Train & Few-shot Eval & Zero-shot Eval & \\
    \midrule
    IIW & 1170 $\times$ 2532 & 1274 & 344 & 87 & 723 & 219 \\
    IIT & 334 $\times$ 720 & 682 & 187 & 45 & 826 & 10 \\
    \bottomrule
  \end{tabular}
\end{table*}

\section{The UI-JEPA Benchmarks}
\label{sec:data}
Current UI understanding benchmarks fall short in capturing UI actions as sequences where temporal relationships can be learned. Additionally, existing datasets focus primarily on learning multimodal representations from separate images, which may not sufficiently capture the complexity of tasks or enable models to accurately predict user intent. These limitations highlight the need for datasets that record task execution as a sequence of UI actions in video format, with each video labeled to describe the user's intent.

For such datasets to be useful practically, they must possess certain characteristics. First, variations in app versions can lead to different presentations and behaviors, and the sequence of UI actions may vary depending on the intended task and app version. Furthermore, to reflect real-world usage and to test model robustness, the dataset should include natural noise, such as irrelevant UI actions in the videos. Regarding challenges, scaling the collection of such a dataset is challenging, requiring a large number of annotators and mobile devices to perform tasks according to instructions, which demands significant investment. Finally, it is impractical to capture all possible visual representation variations of all apps and UI action sequences. To ensure models are built and tested for good generalization, the dataset must support evaluation in both few-shot and zero-shot settings.

To address these challenges, we created two multimodal datasets: ``Intent in the Wild'' which captures open-ended sequences of UI actions with ambiguous user intent, and ``Intent in the Tame'' which focuses on more common tasks with relatively clearer intent. We believe these datasets will contribute to the development of more powerful and lightweight MLLMs, as well as training paradigms with enhanced generalization capabilities. Figure~\ref{fig:data_example} provides examples from each dataset, and Table ~\ref{tab:fata_statistics} provides statistics about each dataset.

\vspace{-1mm}

\subsection{Intent in the Wild}
To train and evaluate our proposed UI-JEPA model, state-of-the-art MLLMs, and other comparable self-supervised learning approaches on tasks where UI actions are open-ended and user intent is challenging to predict or ambiguous, we developed the ``Intent in the Wild'' (IIW) dataset. 
The creation of the IIW dataset involved three key steps:


\begin{itemize}
\item \textbf{Recording UI Interactions}: We manually recorded UI interactions for complex tasks, such as booking a vacation rental, ensuring that some intent categories appear only once. We filtered out any videos shorter than 16 frames to maintain data quality.
\item \textbf{Annotating User Intent}: We annotated the user intent based on the recorded UI interactions, providing high-level, delexicalized descriptions of intent, as illustrated in Figure~\ref{fig:data_example}.
\item \textbf{Dataset Splitting}: The dataset was split into two settings: a few-shot split with at least two instances of videos for each intent category, and a zero-shot split where each intent category appears only once, with no overlap with the few-shot split.
\end{itemize}

The IIW dataset consists of 219 intent categories, with 135 in the few-shot split and 84 in the zero-shot split. In total, the dataset contains 1.6K videos, each averaging 749 frames and approximately $\sim$25 seconds in duration.



\begin{figure}[ht!]
\centering
\begin{subfigure}[b]{0.45\textwidth}
    \centering
    \includegraphics[width=\textwidth]{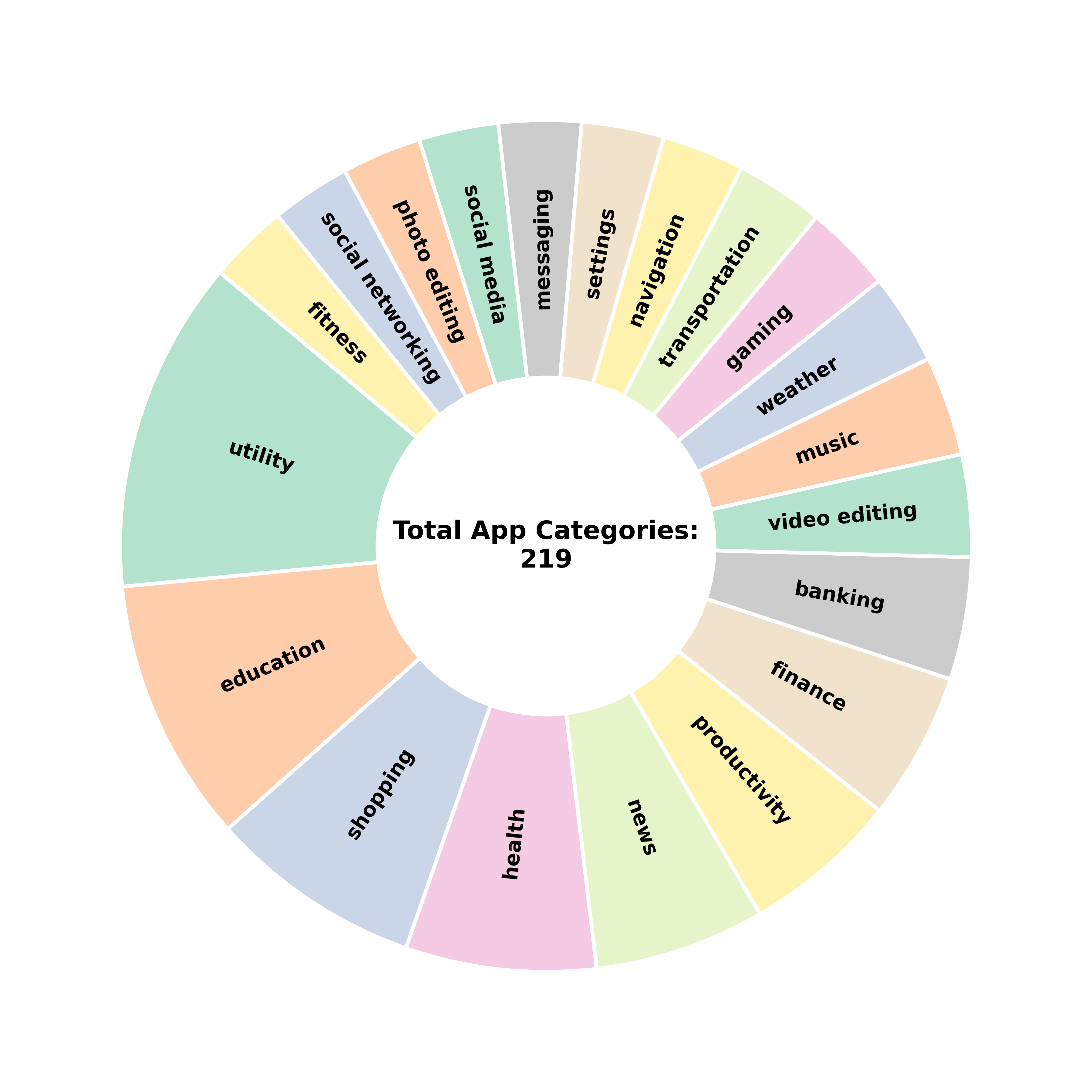}
    \caption{Top 20 most frequent app categories in the IIW Dataset.}
    \label{fig:app_types}
\end{subfigure}
\hfill
\begin{subfigure}[b]{0.45\textwidth}
\centering
\includegraphics[width=\textwidth]{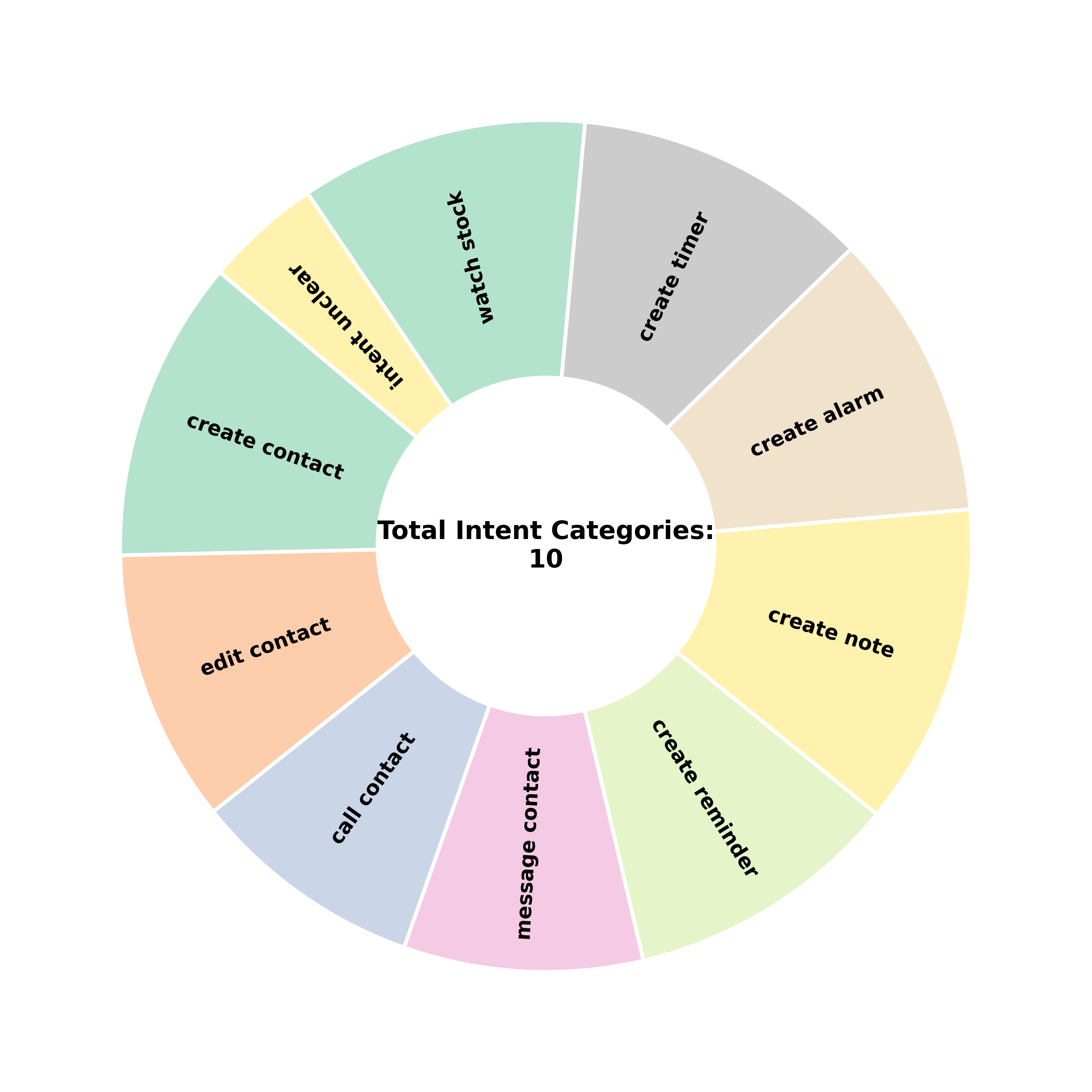}
\caption{10 intent categories in the IIT training split.}
\label{fig:intent_types}
\end{subfigure}
\caption{}
\end{figure}

Figure~\ref{fig:app_types} shows the top 20 most frequent intent categories.

\begin{table*}[ht]
  \caption{Performance Metrics for Different Models on IIW Dataset in the Few-shot Split}
  \label{tab:few_shot_models}
  \begin{tabular}{cccccccccc}
    \toprule
    \textbf{Language Model} & \textbf{Video Encoder} & \textbf{Img. Size} & \textbf{Param.} & \textbf{SBERT} & \textbf{ROUGE-1} &  \textbf{ROUGE-2} & \textbf{ROUGE-L} & \textbf{Intent Sim.} \\
    \midrule
    \multirow{7}{*}{\textbf{Phi}} & Random & 384 & \multirow{7}{*}{\textbf{4.4B}} & 52.46 & 58.10 & 31.94 & 57.48 & 55.94\\
    & MAE & 224 & & 62.82 & 63.77 & 39.33 & 62.97 & 61.87 \\
    & V-JEPA & 384 & & 60.10 & 62.45 & 37.15 & 61.46 & 60.28 \\
    & Random + JEPA tuning & 384 & & 44.02 & 52.37 & 24.12 & 51.78 & 50.07\\
    & MAE + JEPA tuning & 224 & & 59.27 & 61.51 & 36.95 & 60.68 & 59.69 \\
    & UI-JEPA & 224 & & 64.99 & 65.38 & 41.84 & 64.73 & 63.61 \\
    & UI-JEPA & 384 & & \cellcolor{gray!20}66.51 & \cellcolor{gray!20}66.33 & \cellcolor{gray!20}42.94 & \cellcolor{gray!20}65.48 & \cellcolor{gray!20}64.50 \\
    \midrule
    \textbf{Claude 3.5 Sonnet} & --- & Arbitrary & \textbf{>70B} & \textbf{76.58} & 65.12 & 42.06 & 63.58 & 64.76 \\
    \midrule
    \textbf{GPT-4 Turbo} & --- & Arbitrary & \textbf{880B} & 74.79 & 63.91 & 39.61 & 62.54 & 63.36
    \\
    \bottomrule
  \end{tabular}
\end{table*}

\begin{table*}[ht]
  \caption{Performance Metrics for Different Models on IIW Dataset in the Zero-shot Split}
  \label{tab:zero_shot_models}
  \begin{tabular}{cccccccccc}
    \toprule
    \textbf{Language Model} & \textbf{Video Encoder} & \textbf{Img. Size} & \textbf{Param.} & \textbf{SBERT} & \textbf{ROUGE-1} &  \textbf{ROUGE-2} & \textbf{ROUGE-L} & \textbf{Intent Sim.} \\
    \midrule
    \multirow{7}{*}{\textbf{Phi}} & Random & 384 & \multirow{7}{*}{\textbf{4.4B}} & 42.90 & 50.16 & 20.67 & 48.85 & 47.78 \\
    & MAE & 224 & & 49.57 & 53.17 & 22.31 & 51.60 & 50.47 \\
    & V-JEPA & 384 & & 46.82 & 52.48 & 22.93 & 51.21 & 50.01 \\
    & Random + JEPA tuning & 384 & & 41.18 & 49.78 & 20.71 & 49.26 & 47.59 \\
    & MAE + JEPA tuning & 224 & & 47.21 & 52.08 & 22.25 & 50.48 & 49.60 \\
    & UI-JEPA & 224 & & 49.98 & 54.43 & 23.63 & 52.12 & 51.29 \\
    & UI-JEPA & 384 & & \cellcolor{gray!20}50.68 & \cellcolor{gray!20}55.08 & \cellcolor{gray!20}24.71 & \cellcolor{gray!20}53.52 & \cellcolor{gray!20}52.16 \\
    \midrule
    \textbf{Claude 3.5 Sonnet} & --- & Arbitrary & \textbf{>70B} & \textbf{73.74} & \textbf{61.15} & \textbf{34.75} & \textbf{58.63} & \textbf{60.35} \\
    \midrule
    \textbf{GPT-4 Turbo} & --- & Arbitrary & \textbf{880B} & 71.84 & 58.77 & 32.03 & 56.26 & 58.24 \\
    \bottomrule
  \end{tabular}
\end{table*}

\subsection{Intent in the Tame}
The Intent in the Tame (IIT) dataset was developed to record authentic UI interactions executed by individuals in the process of task completion.  For scaling the size, while maintaining realism, dataset creation was automated by framing the navigation as a directed graph \(\langle V, E \rangle\) traversal, where set of vertices \( V \) represents different state of the screen and set of edges \( E \) are the different UI macros (actions). The approach can be summed up in two steps,

\vspace{-2mm}

\begin{itemize}
\item \textbf{Setup}: An iOS framework was used to record diverse set of UI macros across apps. An LLM was utilized to synthetically generate staging data for the device. Macros, in conjunction with the data produced by an LLM, were then used to automatically stage the device to a specific state. For instance, recorded macros for the iOS Contacts app were used to create the contact to be edited (with the name and generated by the LLM) for the Edit Contact intent.
\item \textbf{Execution}: A comprehensive graph was created for each intent which encompassed all potential execution paths. Parameters for the intent were synthetically created via an LLM (to exemplify, the name of the new contact for ``Edit Contact'' intent). Finally, each data point was generated using a staged device and a randomly guided traversal of the graph (algorithm made sure the intent is completed).
\end{itemize}

This approach ensures that the IIT dataset encapsulates four primary characteristics: realism, diversity, intent fulfillment and labeled. In its first version, IIT dataset consists of 914 labeled videos spanning across 10 intent categories as shown in Figure~\ref{fig:intent_types}. Out of 914 videos, 45 videos were evaluated in a zero-shot setting.

\begin{table*}[ht]
  \caption{Performance Metrics for Different Models on IIT Dataset in the Few-shot Split}
  \label{tab:few_shot_models_iit}
  \begin{tabular}{cccccccccc}
    \toprule
    \textbf{OCR}  & \textbf{Language Model} & \textbf{Video Encoder} & \textbf{Param.} & \textbf{SBERT} & \textbf{ROUGE-1} &  \textbf{ROUGE-2} & \textbf{ROUGE-L} & \textbf{Intent Sim.} \\
    \midrule
    \multirow{5}{*}{\textbf{No}} & \multirow{3}{*}{\textbf{Phi}} & Random & \multirow{3}{*}{\textbf{4.4B}} & 59.47 & 58.25 & 39.83 & 55.93 & 58.44 \\
    & & V-JEPA & & 69.69 & 70.40 & 51.72 & 68.06 & 68.76 \\
    & & UI-JEPA & & \cellcolor{gray!20}73.26 & \cellcolor{gray!20}\textbf{73.47} & \cellcolor{gray!20}\textbf{54.86} & \cellcolor{gray!20}\textbf{71.26} & \cellcolor{gray!20}\textbf{71.55} \\
    \cmidrule(l){2-9}
    & \textbf{Claude 3.5 Sonnet} & --- & \textbf{>70B} & \textbf{81.39} & 59.33 & 42.35 & 56.47 & 62.21
 \\
    & \textbf{GPT-4 Turbo} & --- & \textbf{880B} & 79.73 & 59.65 & 36.38 & 55.34 & 60.31 \\
    \midrule
    \multirow{5}{*}{\textbf{Yes}} & \multirow{3}{*}{\textbf{Phi}} & Random & \multirow{3}{*}{\textbf{4.4B}} & 82.68 & 78.78 & 62.38 & 75.84 & 77.09 \\
    & & V-JEPA & & 85.85 & 82.57 & 68.02 & 80.34 & 80.96 \\
    & & UI-JEPA & & \cellcolor{gray!20}\textbf{87.43} & \cellcolor{gray!20}\textbf{83.73} & \cellcolor{gray!20}\textbf{69.17} & \cellcolor{gray!20}\textbf{81.51} & \cellcolor{gray!20}\textbf{82.03} \\
    \cmidrule(l){2-9}
    & \textbf{Claude 3.5 Sonnet} & --- & \textbf{>70B} & 80.67 & 59.05 & 41.78 & 56.63 & 61.95 \\
    & \textbf{GPT-4 Turbo} & --- & \textbf{880B} & 79.29 & 57.88 & 36.65 & 54.61 & 59.70 \\
    \bottomrule
  \end{tabular}
\end{table*}

\begin{table*}[ht]
  \caption{Performance Metrics for Different Models on IIT Dataset in the Zero-shot Split}
  \label{tab:zero_shot_models_iit}
  \begin{tabular}{cccccccccc}
    \toprule
    \textbf{OCR}  & \textbf{Language Model} & \textbf{Video Encoder} & \textbf{Param.} & \textbf{SBERT} & \textbf{ROUGE-1} &  \textbf{ROUGE-2} & \textbf{ROUGE-L} & \textbf{Intent Sim.} \\
    \midrule
    \multirow{3}{*}{\textbf{No}} & \textbf{Phi} & UI-JEPA & \textbf{4.4B} & \cellcolor{gray!20}44.03 & \cellcolor{gray!20}36.12 & \cellcolor{gray!20}12.74 & \cellcolor{gray!20}31.65 & \cellcolor{gray!20}38.13 \\
    \cmidrule(l){2-9}
    & \textbf{Claude 3.5 Sonnet} & --- & \textbf{>70B} & \textbf{79.52} & \textbf{53.21} & \textbf{32.74} & \textbf{49.38} & \textbf{56.27} \\
    & \textbf{GPT-4 Turbo} & --- & \textbf{880B} & 78.26 & 51.75 & 27.43 & 46.43 & 53.69 \\
    \midrule
    \multirow{3}{*}{\textbf{Yes}} & \textbf{Phi} & UI-JEPA & \textbf{4.4B} & \cellcolor{gray!20}41.91 & \cellcolor{gray!20}29.81 & \cellcolor{gray!20}11.66 & \cellcolor{gray!20}27.55 & \cellcolor{gray!20}34.99 \\
    \cmidrule(l){2-9}
    & \textbf{Claude 3.5 Sonnet} & --- & \textbf{>70B} & \textbf{79.69} & \textbf{53.67} & \textbf{33.92} & \textbf{49.85} & \textbf{56.82} \\
    & \textbf{GPT-4 Turbo} & --- & \textbf{880B} & 75.77 & 50.81 & 25.97 & 46.1 & 52.69 \\
    
    \bottomrule
  \end{tabular}
\end{table*}

\section{Baselines}
\label{baselines}
We compare UI-JEPA with several baselines, focusing primarily on different model weights for the video encoder, all based on the ViT architecture~\cite{dosovitskiy2021imageworth16x16words}:
\begin{itemize}
\item Random Encoder: A ViT encoder initialized with random weights, which are then fine-tuned using labeled data in the second stage.
\item V-JEPA: A ViT encoder pre-trained on video datasets using a feature prediction objective.
\item VideoMAE: A ViT encoder pre-trained on video datasets with video tube masking.
\end{itemize}

These baseline video encoders are paired with the same LLM for user intent generation.
We also include closed-source models with potentially different end-to-end architectures for video-to-text generation:
\begin{itemize}
\item GPT-4 Turbo: A closed-source model with multimodal understanding capabilities by OpenAI.
\item Claude-3.5-Sonnet: A closed-source multimodal model by Anthropic.
\end{itemize}
These closed-source models are prompted to convert the UI test split videos into text summaries of user intents for evaluation.

For all baselines, we apply the same data preprocessing technique: sampling 16 evenly spaced frames from the entire video. Since VideoMAE was originally pre-trained with a 224 $\times$ 224 image size, we also report scores for the UI-JEPA encoder using the same 224 $\times$ 224 image size to ensure a fair comparison.

We evaluate the outputs using several established metrics: the SBERT (Sentence-BERT) cosine similarity score~\cite{reimers2019sentencebertsentenceembeddingsusing}, and the ROUGE-1, ROUGE-2, and ROUGE-L scores~\cite{lin-2004-rouge}. The SBERT score measures semantic similarity by embedding sentences into a vector space and computing cosine similarity. ROUGE scores evaluate summary quality through unigram overlap (ROUGE-1), bigram overlap (ROUGE-2), and the longest common subsequence (ROUGE-L), reflecting sentence structure. Additionally, we introduce a new metric, \emph{Intent Similarity}, calculated by averaging the four normalized similarity scores, each scaled to the range [0, 1]. Together, these metrics provide a comprehensive evaluation of both lexical and semantic quality in the generated intents.

\begin{table*}[ht]
  \caption{Performance Metrics for Data Augmentation Techniques}
  \label{tab:data_aug}
  \begin{tabular}{ccccccccc}
    \toprule
    \textbf{Split} & \textbf{Augmentation} & \textbf{SBERT} & \textbf{ROUGE-1} & \textbf{ROUGE-2} & \textbf{ROUGE-L} & \textbf{Intent Sim.} & \textbf{Intent Sim. $\Delta$} \\
    \midrule
    \multirow{4}{*}{\textbf{Few-shot}} & No Augmentation & \cellcolor{gray!20}64.55 & \cellcolor{gray!20}65.42 & \cellcolor{gray!20}41.84 & \cellcolor{gray!20}64.55 & \cellcolor{gray!20}63.52 & \cellcolor{gray!20}0.00\% \\
    & Flipping & \textbf{65.66} & \textbf{65.45} & \textbf{42.06} & \textbf{64.82} & \textbf{63.79} & \textbf{0.42\%} \\
    & Cropping & 63.44 & 64.13 & 40.76 & 63.21 & 62.46 & -1.68\% \\
    & Flipping + Cropping & 61.43 & 62.50 & 37.84 & 61.66 & 60.68 & -4.47\% \\
    \midrule
    \multirow{4}{*}{\textbf{Zero-shot}} & No Augmentation & \cellcolor{gray!20}\textbf{50.68} & \cellcolor{gray!20}\textbf{55.08} & \cellcolor{gray!20}24.71 & \cellcolor{gray!20}\textbf{53.52} & \cellcolor{gray!20}\textbf{52.16} & \cellcolor{gray!20}\textbf{0.00\%} \\
    & Flipping & 47.71 & 52.95 & 22.49 & 51.66 & 50.24 & -3.69\% \\
    & Cropping & 49.59 & 54.33 & \textbf{25.06} & 51.81 & 39.43 & -0.68\% \\
    & Flipping + Cropping & 49.13 & 52.70 & 23.20 & 50.39 & 38.02 & -3.40\% \\
    \bottomrule
  \end{tabular}
\end{table*}

\begin{table*}[ht]
  \caption{Performance Metrics for Adding Positional IDs to Video Embeddings}
  \label{tab:positional_embedding}
  \begin{tabular}{cccccccc}
    \toprule
    \textbf{Split} & \textbf{Add Positional Embeddings} & \textbf{SBERT} & \textbf{ROUGE-1} & \textbf{ROUGE-2} & \textbf{ROUGE-L} & \textbf{Intent Sim.}  \\
    \midrule
    \multirow{2}{*}{\textbf{Few-shot}} & No & \cellcolor{gray!20}\textbf{64.55} & \cellcolor{gray!20}\textbf{65.42} & \cellcolor{gray!20}\textbf{41.84} & \cellcolor{gray!20}\textbf{64.55} &
    \cellcolor{gray!20}\textbf{63.52} \\
    & Yes & 62.95 & 63.64 & 39.61 & 62.65 & 61.84 \\
    \midrule
    \multirow{2}{*}{\textbf{Zero-shot}} & No & \cellcolor{gray!20}\textbf{50.68} & \cellcolor{gray!20}\textbf{55.08} & \cellcolor{gray!20}\textbf{24.71} & \cellcolor{gray!20}\textbf{53.52} &
    \cellcolor{gray!20}\textbf{52.16} \\
    & Yes & 46.89 & 53.08 & 22.64 & 51.44 & 50.15 \\
    \bottomrule
  \end{tabular}
\end{table*}

\begin{figure*}[ht!]
    \centering
    \begin{subfigure}[t]{0.33\textwidth}
        \centering
        \includegraphics[width=\textwidth]{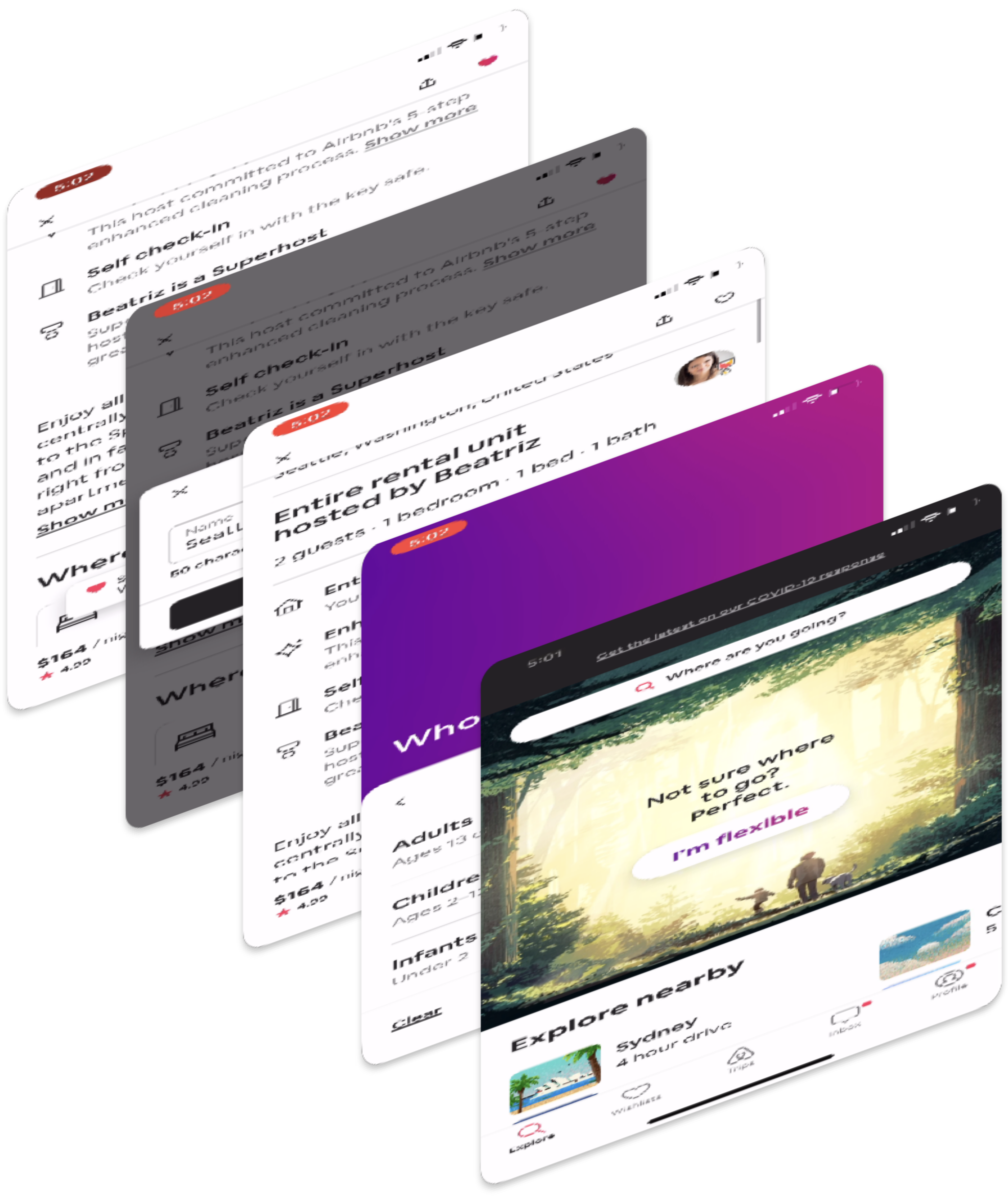}
        \caption{No Masking.}
    \end{subfigure}%
    \hfill
    \begin{subfigure}[t]{0.33\textwidth}
        \centering
        \includegraphics[width=\textwidth]{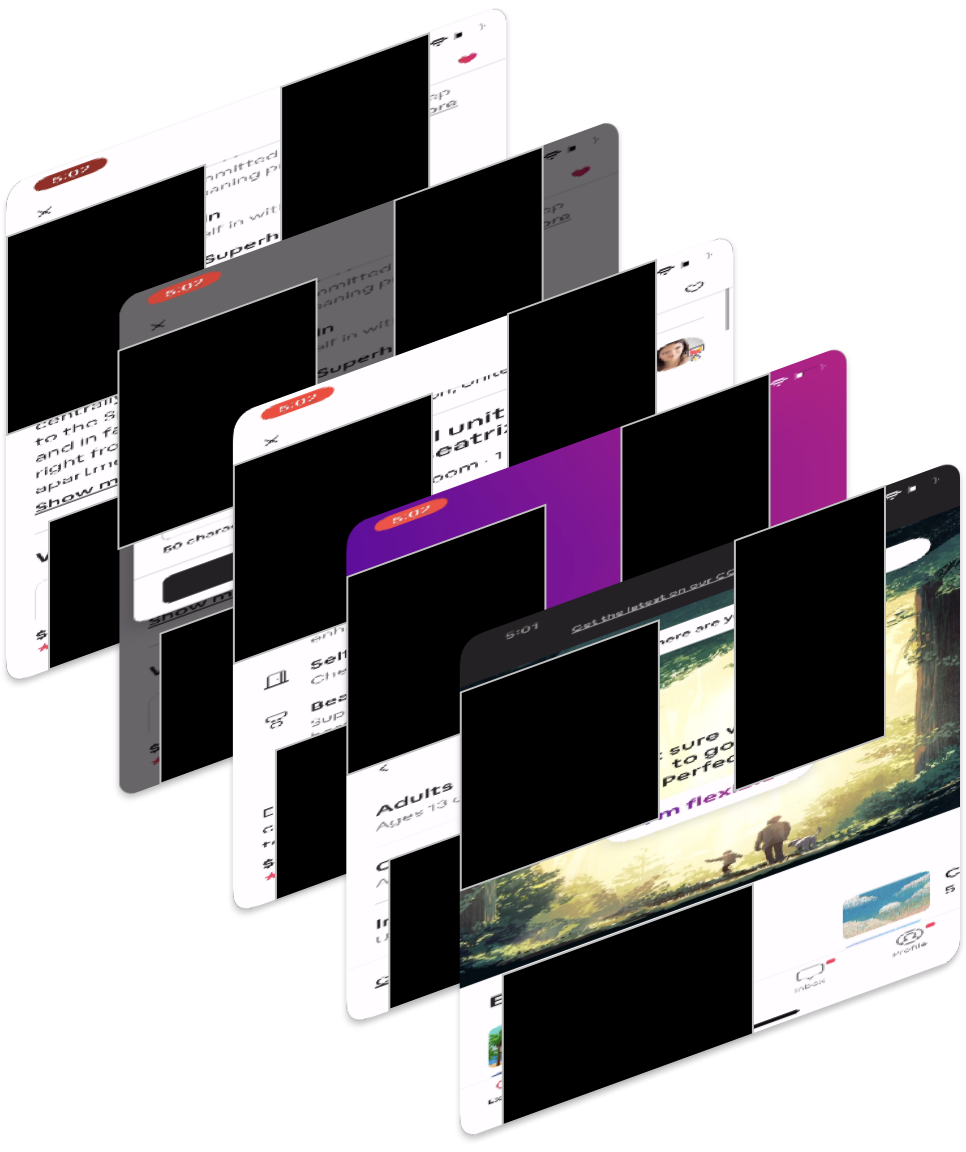}
        \caption{Spatial Masking.}
    \end{subfigure}
    \hfill
    \begin{subfigure}[t]{0.33\textwidth}
        \centering
        \includegraphics[width=\textwidth]{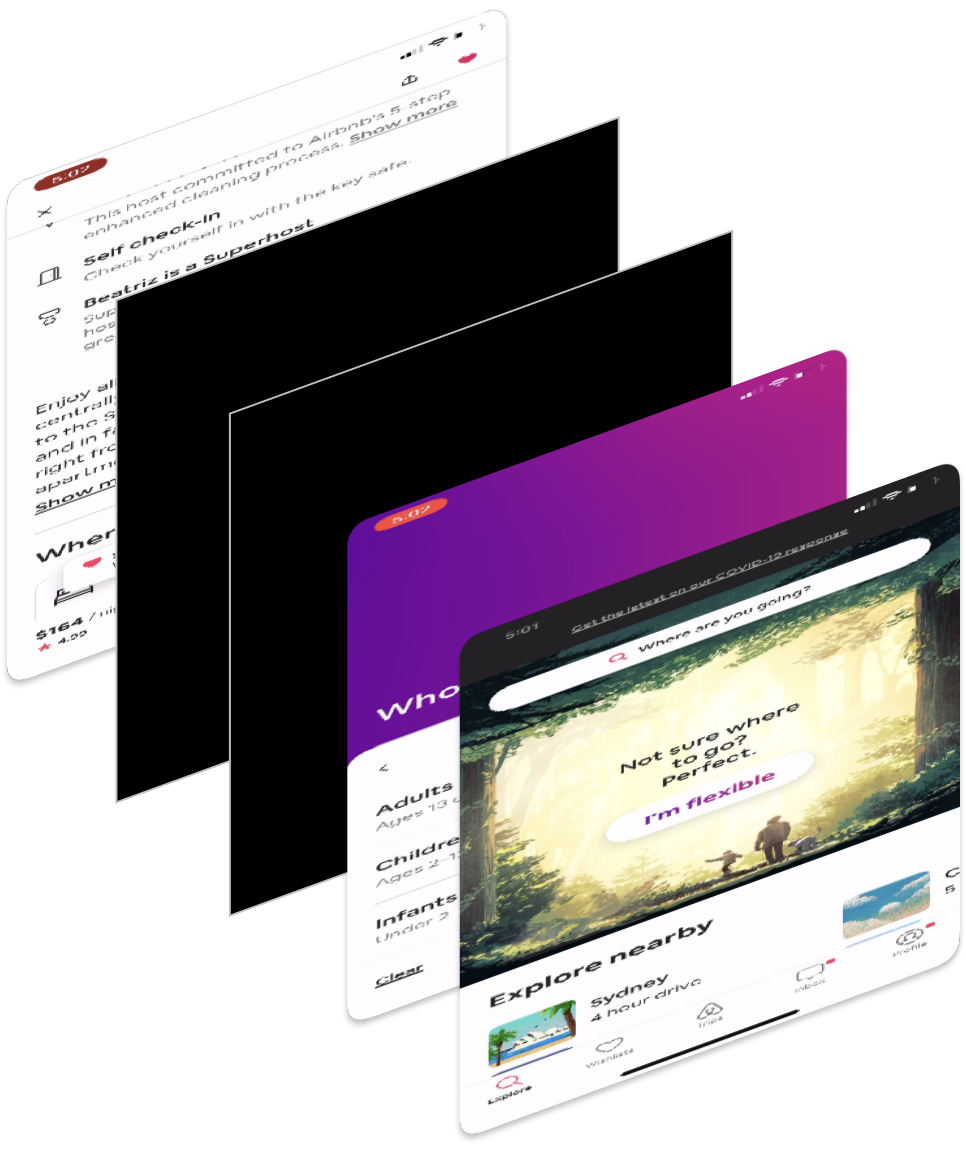}
        \caption{Temporal Masking,}
    \end{subfigure}
    \caption{Example of Masking Strategies: (a) Original Frame Sequences; (b) Consistent Region Masking Across All Frames; (c) Full Region Masking in Selected Frames}
    \label{fig:masking}
\end{figure*}

\section{Results}
\label{sec:results}
\subsection{Main Results}

\paragraph{UI-JEPA outperforms all other baselines on the few-shot split} 

Tables~\ref{tab:few_shot_models}, \ref{tab:zero_shot_models}, \ref{tab:few_shot_models_iit}, and~\ref{tab:zero_shot_models_iit} present the performance of UI-JEPA alongside other baselines on the few-shot and zero-shot splits of the IIW and the IIT datasets, respectively. Among all video encoders, UI-JEPA consistently achieves the highest scores in both few-shot and zero-shot scenarios.
When comparing intent similarity to the leading closed-source model, Claude 3.5 Sonnet, UI-JEPA demonstrates superior performance on few-shot tasks while showing a performance gap in zero-shot settings especially for the IIT dataset. This indicates that while UI-JEPA excels in tasks involving familiar applications, it faces challenges with unfamiliar ones. However, it is important to note that Claude 3.5 Sonnet comes with significant trade-offs, including costs that are 50.5 times higher and latency that is 6.6 times greater than UI-JEPA as a more lightweight model. Incorporating OCR-extracted text from the final frame enhances performance on the few-shot split of IIT dataset for all JEPA-based models, with UI-JEPA outperforming other baselines. However, closed-source models and the zero-shot split show no improvement from this addition. For UI-JEPA specifically, the OCR extraction step introduces a 13.5\% latency increase while delivering a 14.4\% performance boost.

\subsection{Ablation Studies}\label{subsec:abalation}
We perform ablation studies to assess the impact of data augmentation, positional embeddings, masking strategy, masking ratio, and JEPA-tuning data size. All experiments are conducted using the Microsoft Phi-3 model and the "Intent in the Wild" dataset.

\subsubsection{Data Augmentation}
During the pretraining stage of V-JEPA, data augmentation techniques such as random flipping and cropping are applied to each video frame. However, unlike natural video datasets, smartphone screens have a fixed orientation, making image flipping ineffective and potentially introducing noise. Additionally, crucial signals like notifications are often located at the top or bottom of the screen, so cropping risks losing significant information. To evaluate the impact of excluding these less effective data augmentation methods, we conducted experiments (see Table~\ref{tab:data_aug}).

Our results show that while flipping slightly improves performance on the few-shot split, it significantly degrades performance on the zero-shot split. This decline could be due to the video encoder's difficulty in learning a consistent orientation within the UI video dataset, which negatively affects generalization to the zero-shot set. Similarly, the cropping method also leads to reduced performance. The combination of flipping and cropping further exacerbates performance degradation in both few-shot and zero-shot scenarios. As a result, we decided to eliminate all data augmentation methods in our subsequent experiments.

\subsubsection{Positional Embedding}
During the integration of video and text embeddings before passing these hybrid embeddings to the LM, we examined the impact of adding positional embeddings to the video embeddings, similar to those used in text embeddings. We also considered the alternative of omitting these additional positional embeddings. Given that the video embeddings already contain 3D positional information from the encoder, which represents spatio-temporal positions, we evaluated whether adding extra positional embeddings would offer any advantage (see Table~\ref{tab:positional_embedding}).

Our results show that omitting additional positional embeddings consistently improves performance, while their inclusion tends to degrade it. Based on these findings, we chose not to incorporate extra positional embeddings into the video embeddings in our subsequent experiments.

\subsubsection{JEPA-tuning Data Size}
\begin{figure}[ht!]
    \centering
    \includegraphics[width=0.5\textwidth]{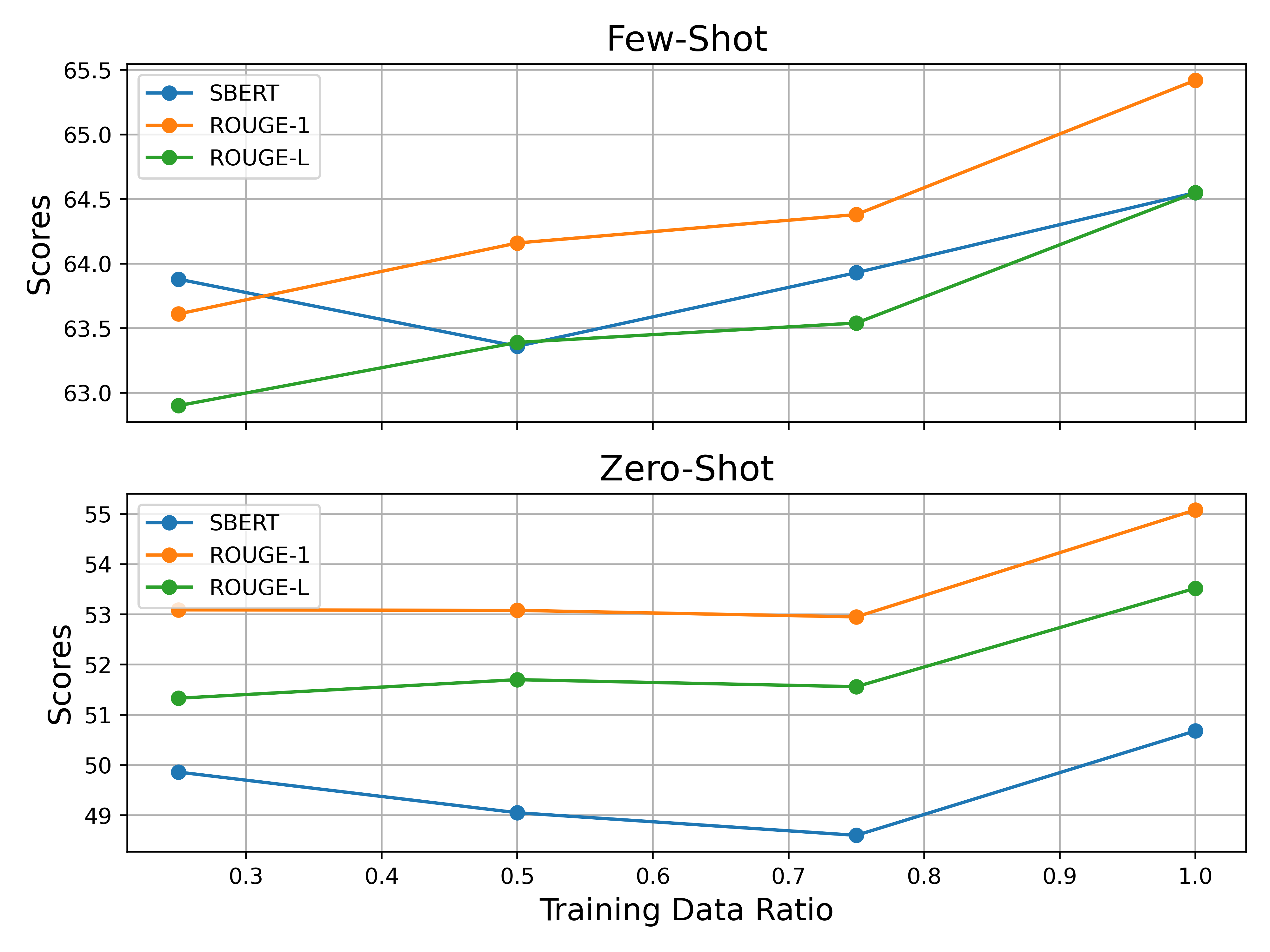}
    \caption{Performance of Different JEPA-tuning Data Size}
    \label{fig:training_ratio}
\end{figure}

In this section, we examine the effect of varying the size of the JEPA-tuning dataset to assess whether adding more unlabeled UI videos enhances video representation and overall model performance. We experimented with using 25\%, 50\%, 75\%, and 100\% of the available data during the JEPA-tuning phase while using the full dataset for fine-tuning. The results are summarized in Figure~\ref{fig:training_ratio}.

As the size of the JEPA-tuning data increases, model performance improves consistently in the few-shot scenario. For the zero-shot scenario, performance initially remains stable but shows a significant increase when the full dataset is used for JEPA-tuning. This indicates that even when the number of labeled examples is fixed, increasing the number of unlabeled examples during JEPA-tuning enhances the model's feature extraction capabilities, leading to better performance in downstream tasks.

\subsubsection{Masking Strategy}
In the original JEPA pre-training process, optimal results are achieved using a multi-block masking strategy, where random spatio-temporal blocks from the entire video are masked. However, applying additional masking to the last few frames can reduce performance. Given the unique characteristics of our UI-grounded multimodal datasets, these conventional masking strategies may not be ideal. To explore the effects of masking during JEPA-tuning, we address two key research questions:

\textbf{Q1:} What types of masking yield optimal performance?

\textbf{Q2:} What ratio and strategy of temporal masking should be used for the best results?

\begin{figure}[ht!]
    \centering
    \includegraphics[width=0.5\textwidth]{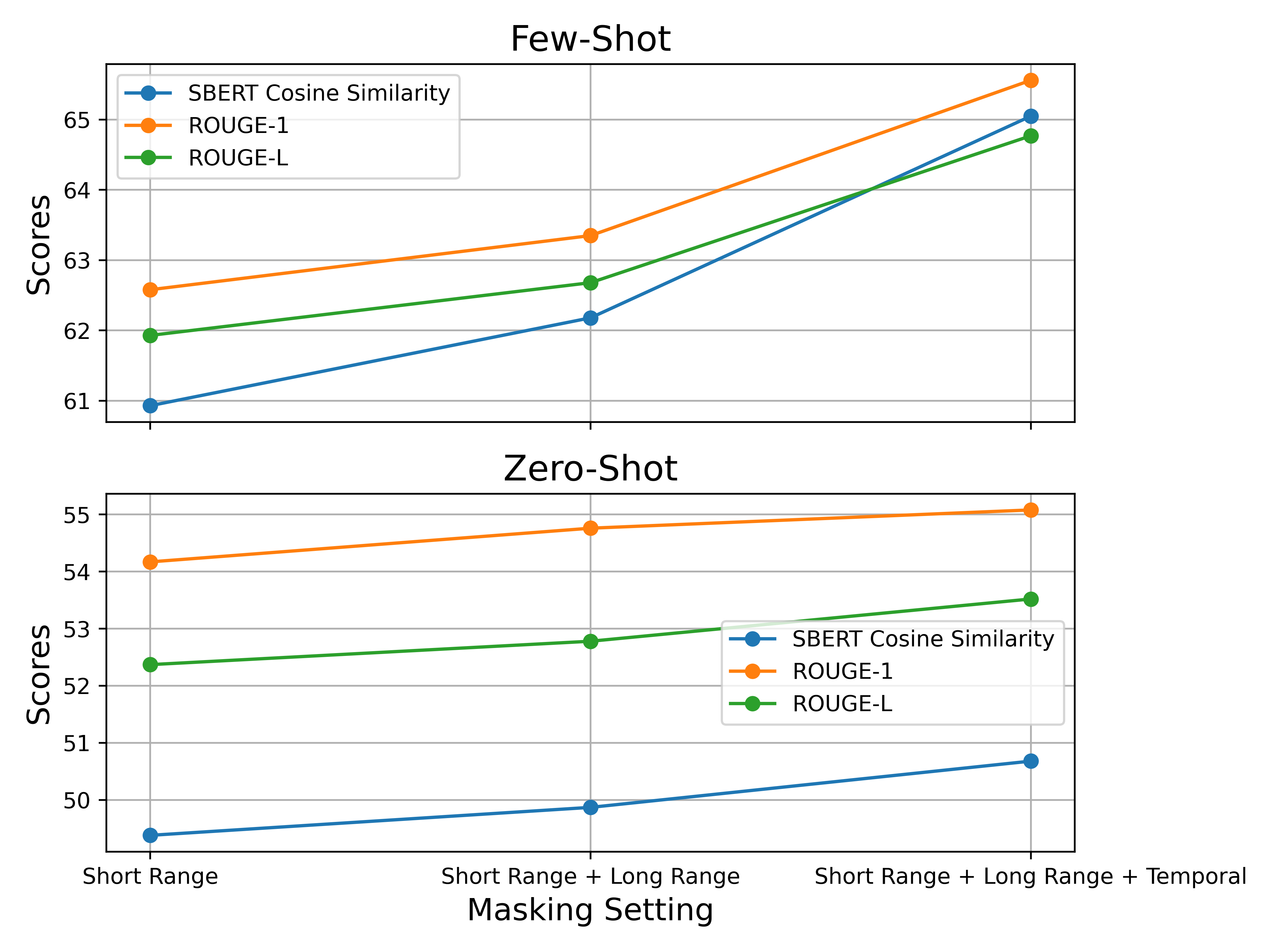}
    \caption{Performance Comparison Across Different Masking Settings}
    \label{fig:masking_type}
\end{figure}

To answer \textbf{Q1}, we experimented during the JEPA-tuning stage with three masking settings: (1) short-range masking, (2) short-range + long-range masking (as used in V-JEPA), and (3) short-range + long-range + temporal masking. Figure~\ref{fig:masking} provides an overview of these masking strategies, and Figure~\ref{fig:masking_type} compares their performance.

\begin{figure*}[ht!]
    \centering
    \includegraphics[width=\textwidth]{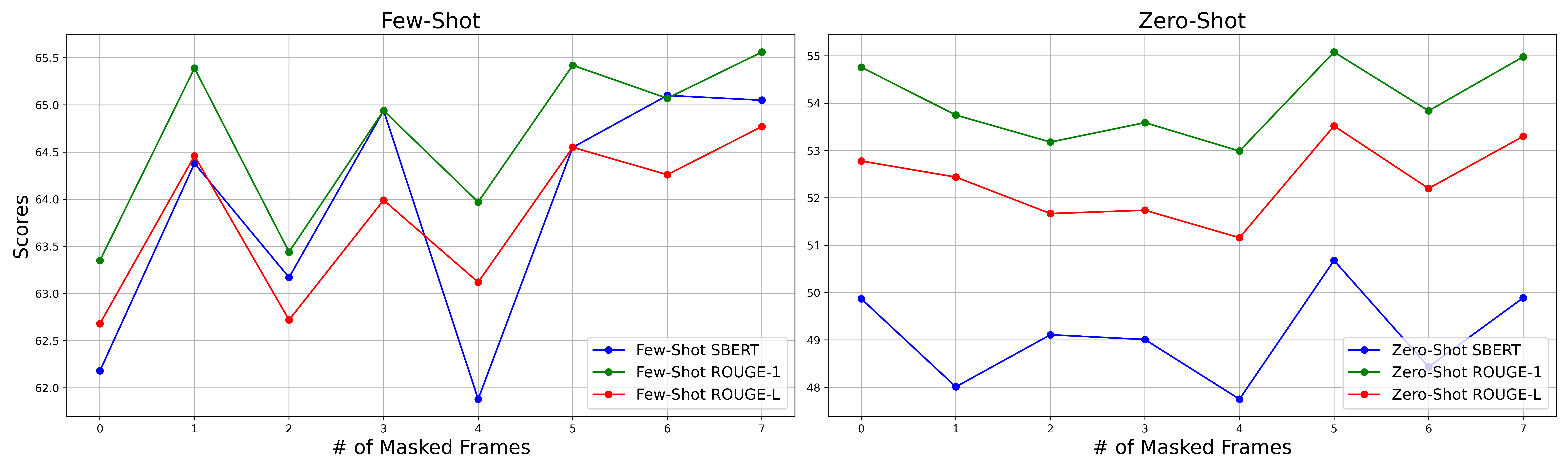}
    \caption{Performance Across Different Numbers of Masked Frames Using Contiguous Temporal Masking.}
    \label{fig:contiguous_masking}
\end{figure*}

\begin{figure*}[ht!]
    \centering
    \includegraphics[width=\textwidth]{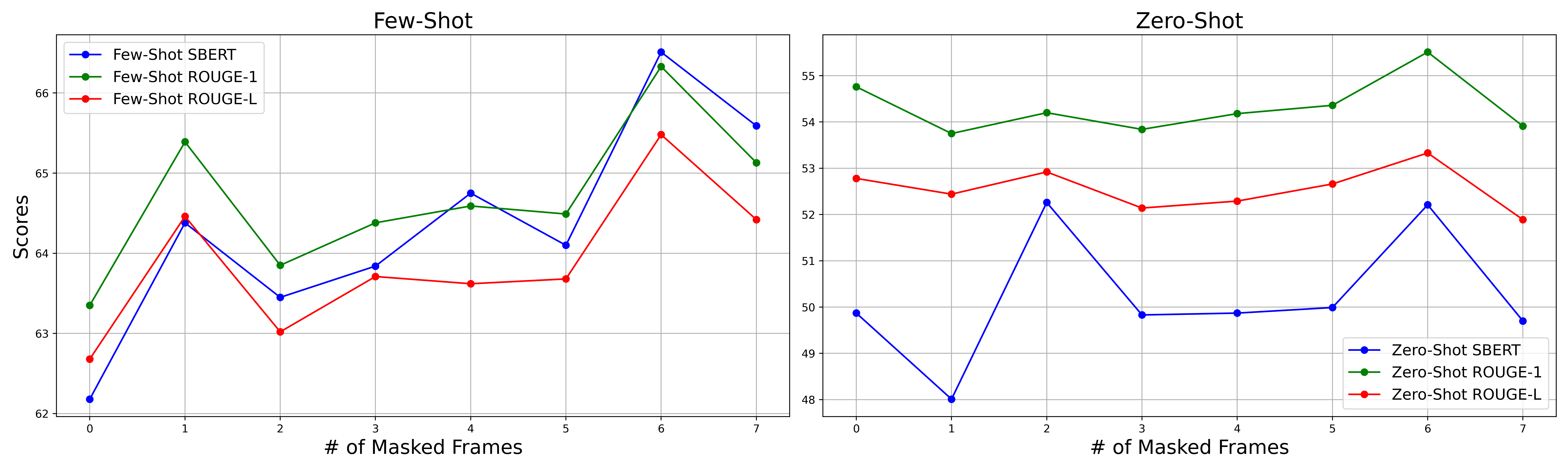}
    \caption{Performance Across Different Numbers of Masked Frames Using Discrete Temporal Masking.}
    \label{fig:discrete_masking}
\end{figure*}

Our results show that progressively adding more masking improves performance in both the few-shot and zero-shot scenarios, indicating that additional temporal masking helps the model better capture serial dependencies between frames, leading to enhanced outcomes.

For \textbf{Q2}, we tested two temporal masking strategies during JEPA-tuning: (1) Contiguous Temporal Masking, where a single block of consecutive frames is masked in Figure~\ref{fig:contiguous_masking}; and (2) Discrete Temporal Masking, where arbitrary frames are selected for masking in Figure ~\ref{fig:discrete_masking}. Since each patch spans two video frames, masking occurs at the level of these patches, with ``frame'' here referring to a hyper-frame consisting of two video frames.


From our experiments, we observed that in the few-shot scenario, performance improves as more frames are masked. However, in the zero-shot scenario, this trend is less pronounced, with performance remaining relatively stable across different masking ratios. The best results for both the few-shot and zero-shot scenarios are achieved with six masked discrete frames.

\vspace{-3mm}

\begin{figure*}[ht!]
    \centering
    \includegraphics[width=\textwidth]{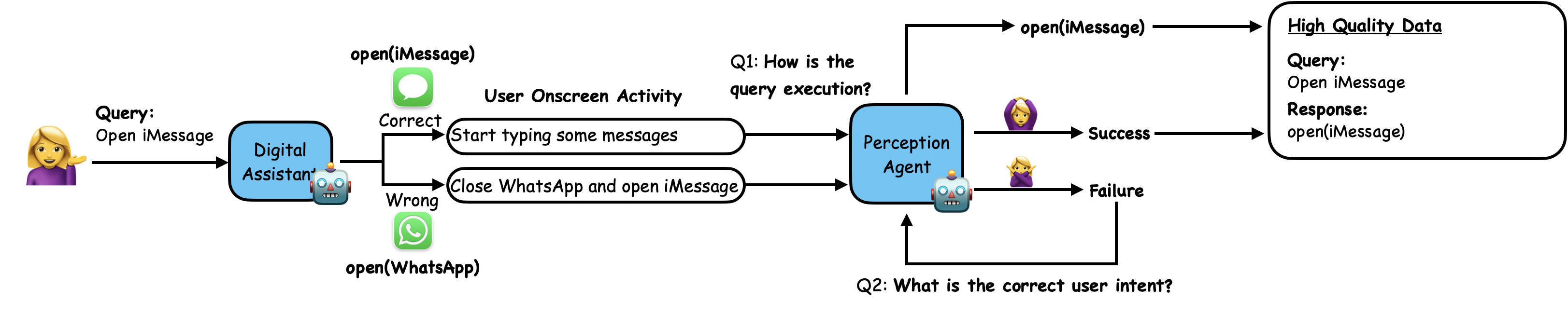}
    \caption{User Feedback Learning: UI-JEPA automatically filters and labels high-quality data for constructing datasets to train digital assistants.}
    \label{fig:ufl}
\end{figure*}

\section{Conclusion and Applications}
\label{conclusion}

\subsection{Conclusion}
In this work, we introduced \textbf{UI-JEPA}, a framework that uses self-supervised learning to generate abstract UI embeddings and, when combined with a small LLM, can perform user intent prediction. UI-JEPA matches the performance of state-of-the-art MLLMs while reducing computational costs by 50.5x and latency by 6.6x, making it ideal for lightweight, on-device applications. Our newly introduced datasets,``Intent in the Wild'' (IIW) and ``Intent in the Tame'' (IIT), establish a benchmark for few-shot and zero-shot UI understanding. These findings highlight UI-JEPA's potential for advancing efficient and privacy-preserving UI understanding.

\subsection{Applications of UI-JEPA}
User intent understanding has at least two key applications: User Feedback Learning and Multimodal Intent State Tracking.

\subsubsection{User Feedback Learning}
Smartphone users worldwide interact daily with digital assistants, generating a vast array of queries. This data is invaluable for refining the reasoning capabilities of digital assistants and aligning their responses with user preferences. However, privacy and security concerns aside, a significant challenge is that many digital assistants currently struggle to address user requests effectively, resulting in a large volume of disorganized, low-quality data. By using UI-JEPA, we can accurately infer user intent from on-screen activity and assess the effectiveness of the digital assistant's performance. This includes determining whether users continue with the app opened by the assistant or switch to a different one. If UI-JEPA predicts successful execution, the data point can be directly added to our high-quality dataset. Conversely, if the prediction indicates a failure, UI-JEPA can still capture the user’s true intent, contributing valuable data to our high-quality dataset, as shown in Figure~\ref{fig:ufl}. These high-quality datasets can then be used to further fine-tune UI-JEPA, enhancing its performance and enriching the dataset for future UI understanding research.

\subsubsection{Multimodal Intent State Tracking}
\begin{figure}[ht!]
    \centering
    \includegraphics[width=\columnwidth]{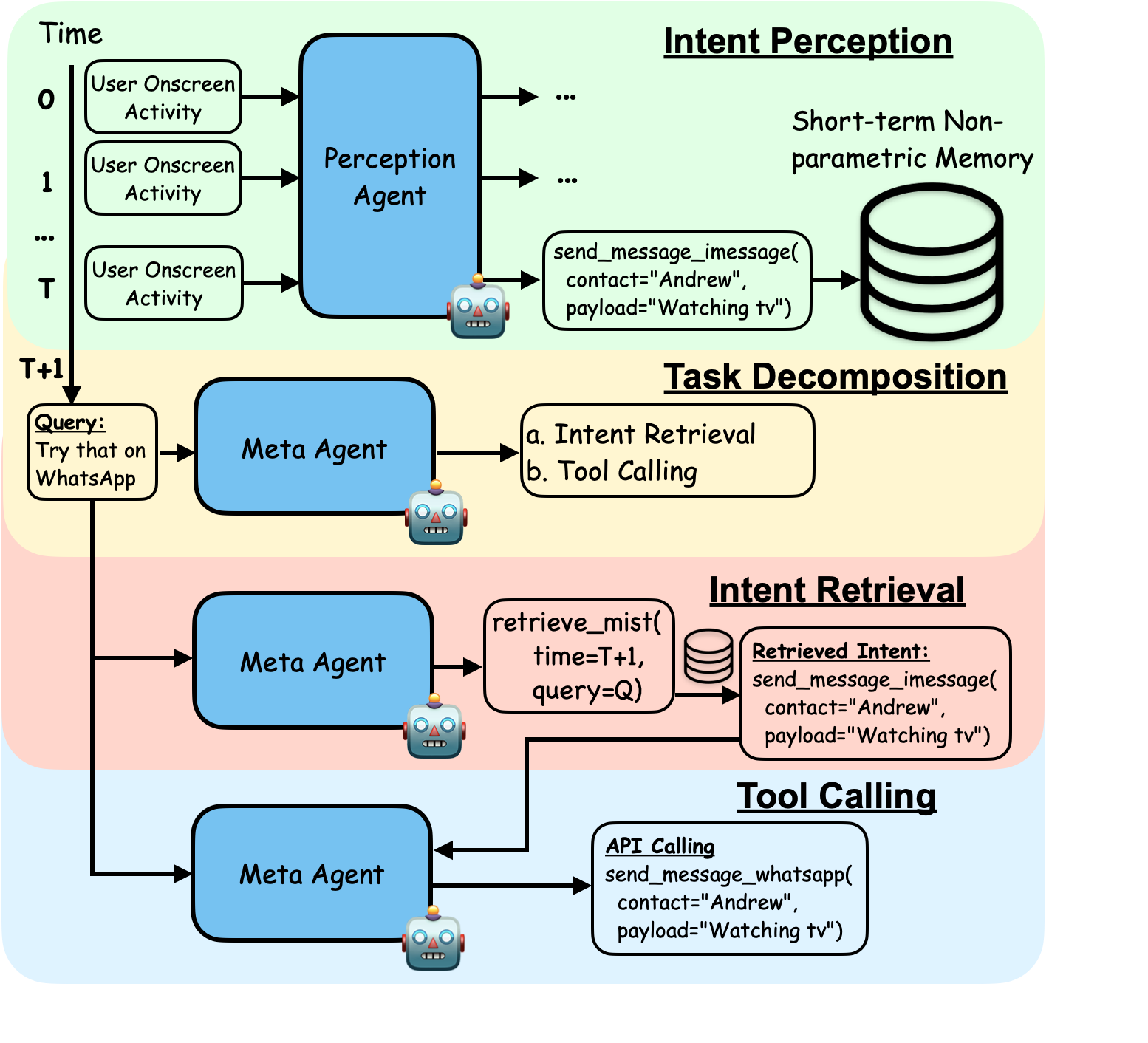}
    \caption{Multi-modal Intent State Tracking: This framework integrates UI-JEPA as a perception agent that actively monitors and captures user intent across various applications and modalities.}
    \label{fig:mist}
\end{figure}

Another promising application of UI-JEPA is its integration into an agentic framework \cite{qiao-etal-2024-autoact, suzgun2024metapromptingenhancinglanguagemodels} designed to actively track user intent across various applications and modalities. In this framework, UI-JEPA functions as the perception agent, capturing user intent at different time points and storing these intents in a memory store. When a user interacts with a digital assistant, the system retrieves the most relevant intent based on the query and generates the appropriate API call to fulfill the user's request, as illustrated in Figure~\ref{fig:mist}.

\section{Limitations}
\label{limitations}
While UI-JEPA has shown promising results on the IIW and IIT datasets, several limitations remain:

\begin{itemize}
    \item Granularity of User Intent Prediction: JEPA embeddings alone often fall short for granular user intent prediction, especially in the IIT dataset, where the encoder primarily captures high-level video representations. This limitation affects tasks requiring detailed text recognition and description. To address this, we incorporate OCR; however, its effectiveness is contingent on the quality of the OCR and the presence of textual information in the frames. Further research is needed to enhance JEPA representations to capture more detailed content.
    
    \item Pre-training Requirements: Experimental results reveal that JEPA-tuning a randomly initialized video encoder yields poor performance. Effective JEPA-tuning necessitates extensive pre-training of the video encoder, which restricts the applicability of JEPA-tuning to scenarios with ample video data.
    
    \item Performance in Zero-Shot Scenarios: Although UI-JEPA performs competitively with large MLLMs like Claude 3.5 Sonnet and GPT-4 Turbo in few-shot settings, its performance lags significantly in zero-shot scenarios. This indicates that UI-JEPA's ability to generalize from familiar to unfamiliar apps needs improvement.
    
    \item Audio Modality: UI-JEPA has not been tested with audio modalities, and its performance in this domain remains unexamined.
\end{itemize}

\begin{acks}
We would like to thank Amanda Swearngin, Barry Theobald, Chen Huang, Eldon Schoop, Arturo Argueta, and Vimal Thilak for their insightful feedback and comments on the draft. We are also grateful to Srinivas Chappidi, Sophie Ostlund, Luisa Mendoza Gonzalez, Gabriela Lopez, Josh Elman, Tim Kolecke, Diane White, Steve Smith, Lindsay Hislop, and John Giannandrea for their invaluable support throughout this work.
\end{acks}

\bibliographystyle{ACM-Reference-Format}
\bibliography{main}

\appendix

\section{Dataset Processing Details}
In the V-JEPA paper, encoders process video clips consisting of 16 frames, with a temporal stride of 4 between consecutive frames. For the K400 dataset, 8 clips are sampled per video, while for the SSv2 dataset, only 1 clip is sampled per video. However, this approach relies on prior knowledge of video distribution and may not be effective for videos of varying lengths. To address this, we propose sampling a single 16-frame clip using a flexible temporal stride, with frames evenly sampled from the beginning to the end of the video. This method accommodates variations in video length more effectively.

During the JEPA-tuning and fine-tuning stages, we exclude videos with fewer than 16 frames, as such brief videos are unlikely to capture significant user activities.

\section{Training Details}
In this section, we provide details on the training process for UI-JEPA, including both the JEPA-tuning and fine-tuning stages. Table~\ref{tab:hyperparameters} summarizes the key hyperparameters used during pretraining.

\begin{table*}[ht]
\centering
\caption{Hyper-parameters and Their Values}
\label{tab:hyperparameters}
\begin{tabular}{lcccc}
\toprule
\multirow{2}{*}{\textbf{Hyper-parameter}} & \multicolumn{2}{c}{\textbf{IIW}} & \multicolumn{2}{c}{\textbf{IIT}} \\
\cmidrule(l){2-5}
& \textbf{JEPA-tuning} & \textbf{Fine-tuning} & \textbf{JEPA-tuning} & \textbf{Fine-tuning} \\
\midrule
\textbf{Data} & & & & \\
Resolution & 384 & 384 & 384 & 384 \\
Num Frames & 16 & 16 & 16 & 16 \\
Temporal Stride & Flexible & Flexible & Flexible & Flexible \\
Data Augmentation & False & False & False & False \\
\midrule
\textbf{Short-Range Masking} & & & & \\
Block Aspect Ratio & (0.75, 1.5) & ---  & (0.75, 1.5) & --- \\
Num Blocks & 8 & --- & 8 & --- \\
Spatial Scale & 0.15 & --- & 0.15 & --- \\
Temporal Scale & 1 & --- & 1 & --- \\
\midrule
\textbf{Long-Range Masking} & & \\
Block Aspect Ratio & (0.75, 1.5) & --- & (0.75, 1.5) & --- \\
Num Blocks & 2 & --- & 2 & --- \\
Spatial Scale & 0.7 & --- & 0.7 & --- \\
Temporal Scale & 1 & --- & 1 & --- \\
\midrule
\textbf{Temporal Masking} & & & & \\
Block Aspect Ratio & (1.0, 1.0) & --- & (1.0, 1.0) & --- \\
Num Blocks & 1 & --- & 1 & --- \\
Spatial Scale & 1 & --- & 1 & --- \\
Temporal Scale & 0.75 & --- & 0.75 & --- \\
\midrule
\textbf{Optimization} & & & & \\
Batch Size & 4 & 1 & 4 & 1 \\
Total Number of Iterations & 4000 & 6000 & 2000 & 3000 \\
Warmup Iterations & 100 & 300 & 50 & 150  \\
LR & $3\times10^{-4}$ & $3\times10^{-4}$ & $3\times10^{-4}$ & $3\times10^{-4}$  \\
Start LR & $2\times10^{-4}$ & $2\times10^{-4}$ & $2\times10^{-4}$ & $2\times10^{-4}$  \\
Final LR & $1\times10^{-6}$ & $1\times10^{-6}$ & $1\times10^{-6}$ & $1\times10^{-6}$  \\
Start Momentum & 0.998 & 0.998 & 0.998 & 0.998 \\
Final Momentum & 1.0 & 1.0 & 1.0 & 1.0 \\
Start Weight Decay & 0.04 & 0.04 & 0.04 & 0.04 \\
Final Weight Decay & 0.4 & 0.4 & 0.4 & 0.4 \\
Scheduler Scale Factor & 1.25 & 1.25 & 1.25 & 1.25 \\
\midrule
\textbf{Architecture} & & & & \\
Patch Size & 16 & 16 & 16 & 16 \\
Tubelet Size & 2 & 2  & 2 & 2 \\
Pred Depth & 12 & 12 & 12 & 12 \\
Pred Embed Dim & 12 & 12 & 12 & 12 \\
\midrule
\textbf{Hardware} & &  & &  \\
Data Type & bfloat16 & bfloat16 & bfloat16 & bfloat16 \\
Accelerator & A100 80G & A100 80G & A100 80G & A100 80G \\
\bottomrule
\end{tabular}
\end{table*}

\subsection{Fine-tuning}

\subsubsection{Separator and Ending Token}
During fine-tuning, we insert a separator token between the video embedding and the OCR text, as well as between the OCR text and the user intent. An ending token is also used to distinguish between different input types. For the Microsoft Phi-3 model, the separator token is <|placeholder1|>, and the ending token is <|endoftext|>.

\subsubsection{Positional Embedding}
For positional embeddings, we do not apply additional positional embeddings to the video embedding or the initial separator token. Instead, we use the standard positional ID starting from the first OCR text token. If no OCR text is present, the positional ID begins with the first intent text token.

\subsubsection{OCR}
In our study, we use the Apple Vision Service for Optical Character Recognition (OCR), employing the VNRecognizeTextRequest to extract text from images. To improve the relevance of the extracted data, we filter out single-letter texts and specific strings such as "123", "space", and "return." These elements generally correspond to keyboard inputs displayed on the user's screen and do not provide meaningful information.

\subsubsection{LoRA Configuration}
We apply the LoRA tuning technique during the fine-tuning of the LM. For detailed information on the LoRA tuning configuration, please refer to Table~\ref{tab:lora}.

\begin{table*}[ht]
\centering
\caption{LoRA Tuning Configuration}
\label{tab:lora}
\begin{tabular}{lc}
\toprule
\textbf{Hyper-parameter} & \textbf{Parameter Value} \\
\midrule
\textbf{LoRA} & \\
LoRA Alpha & 16 \\
LoRA Dropout & 0.05 \\
LoRA Rank & 16 \\
Target Modules & $qkv\_proj, o\_proj, gate\_up\_proj, down\_proj$ \\
\midrule
\textbf{Quantization} & \\
Quantization Type & nf4 \\
Double Quantization & true \\
Computation Type & bfloat16 \\
\bottomrule
\end{tabular}
\end{table*}

\section{Inference}

\subsection{Prompts}
For closed-source MLLMs, including GPT-4 Turbo and Claude 3.5 Sonnet, the prompts used to generate results are detailed in Table~\ref{tab:prompt}.

\begin{table*}[ht]
\centering
\caption{Prompts used for baselines}
\label{tab:prompt}
\begin{tabular}{p{\textwidth}}
\toprule
\textbf{Prompt} \\
\midrule
\textbf{Prompt for IIW Dataset:} \\

Here is a recording of a user's operation on an iPhone. Please summarize the user's intent in a delexicalized and concise manner, removing specific text detail, and specify the app type following the format below using plain text: \\ \\

Intent:

App Type: \\ \\
            
Here are some examples: \\ \\

Intent: The user checks weather.

App Type: weather \\ \\

Intent: The user browses products.

App Type: shopping \\ \\

Intent: The user looks for accommodations.

App Type: travel \\ \\

Here is the recording:

\{image1\}\{image2\}, ..., \{image16\}. \\
\midrule
\textbf{Prompt for IIT Dataset:} \\

Here is a recording of a user's iPhone activity. Please summarize their intentions. While the intent may involve multiple activities, it must end with one of the following activity categories: \\ \\

Example Ending Activity Categories: \\
User calls Abigail from their contacts. \\
User updates a contact name to Alex. \\
User sends a message to Andrew via the iMessage app saying 'Watching TV.'. \\
User creates an alarm for 11:09 AM. \\
User adds Mastercard stock to their watchlist. \\
User adds a new contact named Ravi. \\
User adds a reminder 'Doctor appointment'. \\
User creates a note titled 'Presentation notes' within the 'Resolutions' folder. \\
User creates a timer for 6 minutes and 52 seconds. \\ \\

Example output: \\
User opens stock app, and then calls Lily from their category. \\
User enables Do Not Disturb, searches for contacts, and adds a new contact named Jackson. \\
User searches for Netflix, opens app, and creates an alarm for 2:19 AM. \\ \\

Here is the recording:

\{image1\}\{image2\}, ..., \{image16\}. \\  \\

(Optional) Here is the OCR text from the final frame: \{OCR\_text\}. \\

\bottomrule
\end{tabular}
\end{table*}

When the byte size of 16 images in the Claude 3.5 Sonnet dataset exceeds processing limits, we use only half of the images (specifically, images 1, 3, 5, ..., 15) to obtain results. Additionally, if Claude 3.5 Sonnet fails to provide descriptions of user intent or application type due to ethnic or privacy considerations, we manually exclude these responses from the performance metrics calculation.

\end{document}